\documentclass[sn-apa, iicol, pdflatex]{arxiv}

\usepackage{bm}

\jyear{2023}%

\theoremstyle{thmstyleone}%
\theoremstyle{thmstyletwo}%

\theoremstyle{thmstylethree}%

\begin{document}

\title[MMOT]{MMoT: Mixture-of-Modality-Tokens Transformer for Composed Multimodal Conditional Image Synthesis}


\author[1]{\fnm{Jianbin} \sur{Zheng}}\email{jabir.zheng@outlook.com}
\equalcont{These authors contributed equally to this work.}

\author[2]{\fnm{Daqing} \sur{Liu}}\email{liudq.ustc@gmail.com}
\equalcont{These authors contributed equally to this work.}

\author*[2]{\fnm{Chaoyue} \sur{Wang}}\email{chaoyue.wang@outlook.com}

\author[3]{\fnm{Minghui}\sur{Hu}}\email{e200008@e.ntu.edu.sg}

\author[4]{\fnm{Zuopeng} \sur{Yang}}\email{yzpeng@sjtu.edu.cn}

\author*[1]{\fnm{Changxing} \sur{Ding}}\email{chxding@scut.edu.cn}

\author[2]{\fnm{Dacheng} \sur{Tao}}\email{dacheng.tao@gmail.com}

\affil*[1]{\orgdiv{School of Electronic and Information Engineering}, \orgname{South China University of Technology}, \orgaddress{\city{Guangzhou}, \postcode{510006}, \country{China}}}

\affil*[2]{\orgname{JD Explore Academy}, \orgaddress{\city{Beijing}, \postcode{100176}, \country{China}}}

\affil[3]{\orgdiv{School of Electrical and Electronic Engineering}, \orgname{Nanyang Technological University}, \orgaddress{\city{Singapore}, \postcode{639798}, \country{Singapore}}}

\affil[4]{\orgdiv{Department of Automation}, \orgname{Shanghai Jiao Tong University}, \orgaddress{\city{Shanghai}, \postcode{200240}, \country{China}}}

\abstract{Existing multimodal conditional image synthesis (MCIS) methods generate images conditioned on any combinations of various modalities that require all of them must be exactly conformed, hindering the synthesis controllability and leaving the potential of cross-modality under-exploited.
To this end, we propose to generate images conditioned on the compositions of multimodal control signals, where modalities are imperfectly complementary, \textit{i.e.}, composed multimodal conditional image synthesis (CMCIS).
Specifically, we observe two challenging issues of the proposed CMCIS task, \textit{i.e.}, the modality coordination problem and the modality imbalance problem.
To tackle these issues, 
we introduce a Mixture-of-Modality-Tokens Transformer (MMoT) that adaptively fuses fine-grained multimodal control signals,
a multimodal balanced training loss to stabilize the optimization of each modality, and a multimodal sampling guidance to balance the strength of each modality control signal.
Comprehensive experimental results demonstrate that MMoT achieves superior performance on both unimodal conditional image synthesis (UCIS) and MCIS tasks with high-quality and faithful image synthesis on complex multimodal conditions. The project website is available at \url{https://jabir-zheng.github.io/MMoT}.}




\maketitle

\section{Introduction}
\begin{figure}
	\centering
	\includegraphics[width=0.95\linewidth]{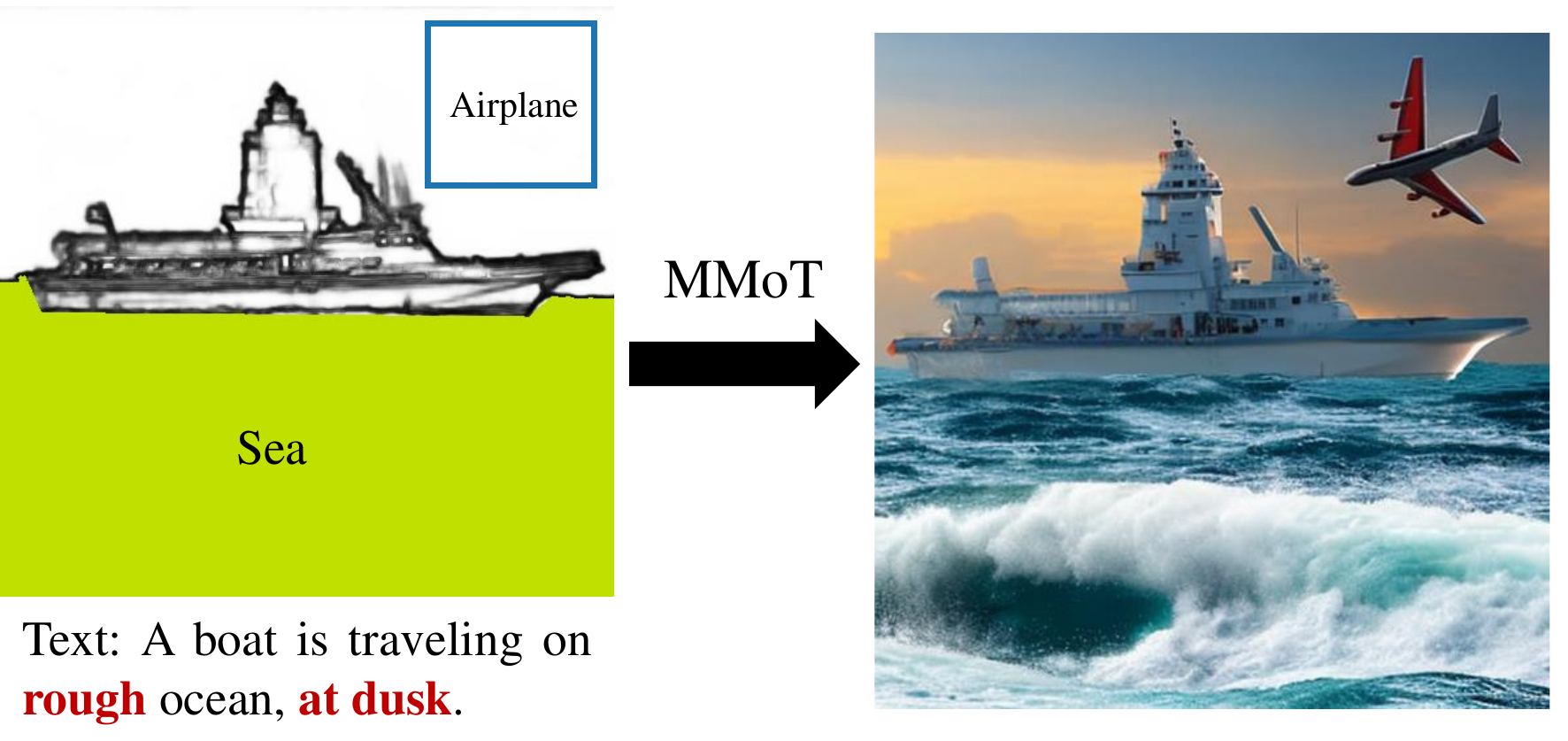}
	\caption{CMCIS relaxes the stringent requirements for inputs. The input can be the composition of multiple complementary modalities (\textit{e.g.}, text, sketch, segmentation mask, and bounding boxes), and our MMoT can generate reasonable results leveraging all inputs.}
	\label{fig:cmcis}
\end{figure}

\begin{figure*}
    \centering
    \includegraphics[width=\linewidth]{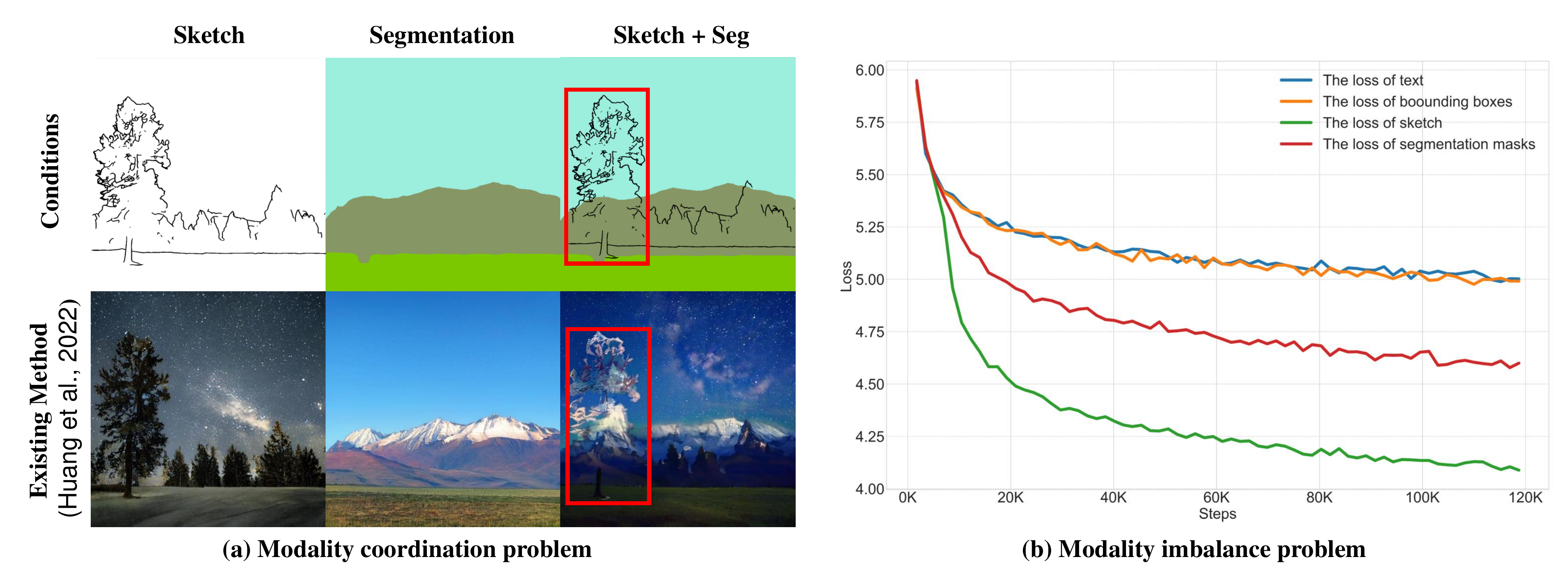}
    \caption{ (a) Modality coordination problem: incorrect coordination of multiple modalities, \textit{e.g.}, the tree incorrectly composed with the mountain. (b) Modality imbalance problem: various modalities tend to converge at different rates and converge to different endpoints.}
    \label{fig:problem}
\end{figure*}


With the maturity of image synthesis quality, we start focusing on improving its controllability to generate specific images as we expected.
To control the image content, various control signals from different modalities have been proposed, including texts~\citep{zhou2021lafite}, sketches~\citep{wang2018high}, segmentation masks~\citep{wang2018high}, and bounding box layouts~\citep{sun2021learning}, where each of them has its own advantages, \textit{e.g.}, texts usually describe attributes of objects and style of images, segmentation masks depict the environmental contexts, and sketches/layouts are able to control the position and size of each object.

The pioneering methods mostly focus on unimodal conditional image synthesis (\textbf{UCIS}) that generates images with a single unimodal control signal, \textit{e.g.}, text-to-image~\citep{zhou2021lafite,ramesh2021zero,ramesh2022hierarchical,saharia2022photorealistic,yu2022scaling}, sketch-to-image~\citep{wang2018high,park2019semantic,esser2021taming,wang2022pretraining}, segmentation-to-image~\citep{wang2018high,park2019semantic,sushko2022oasis,esser2021taming,wang2022pretraining}, and layout-to-image~\citep{zhao2020layout2image,sun2021learning,sylvain2021object,li2021image,he2021context,yang2022modeling}.
Despite the great progress they have achieved, those methods fail to fully utilize the information from different modalities, thus hindering their controllability and applications in real scenarios.
To this end, recent works~\citep{zhang2021ufc,li2022m6,huang2022multimodal} propose to generate images conditioned on any combinations of various modalities, \textit{e.g.}, sketch+segmentation or text+segmentation, termed multimodal conditional image synthesis (\textbf{MCIS}).
However, the current MCIS strictly requires every unimodal signal must be \textit{exactly} conformed with each other, leaving the potential of cross-modality under-exploited.
Additionally, it is unfriendly to the majority of users without a professional painting background. 

In this paper, we propose a more challenging task, namely Composed Multimodal Conditional Image Synthesis (\textbf{CMCIS}), which allows each unimodal signal to be imperfectly complementary and the final expectation of images to be composed of all modality signals.
For example, as illustrated in Figure~\ref{fig:cmcis}, the user can generate a desired image by using a variety of different modalities to describe different components of the scene, \textit{i.e.}, drawing a sketch of the boat, filling the sea with a segmentation mask, using a bounding box to decide the size and location of the airplane, and the text giving high-level semantic information such as dusk.

Unfortunately, existing MCIS methods, no matter GAN-based~\citep{huang2022multimodal}, Transformer-based~\citep{zhang2021ufc, li2022m6}, or diffusion-based~\citep{huang2023composer}, are not able to handle the new challenging CMCIS task.
As illustrated in Figure~\ref{fig:problem}, we observe two main issues of existing MCIS methods: \textbf{(i)} the modality coordination problem due to the nonadaptive fusion on fine-grained information across multiple modalities, \textit{e.g.}, the tree incorrectly composed with the texture of the mountain in Figure~\ref{fig:problem} (a); and \textbf{(ii)} the modality imbalance problem that caused by the imbalanced distribution of each modality in datasets, \textit{i.e.}, different modalities tend to converge at different rates in Figure~\ref{fig:problem} (b).

Specifically, the \textit{modality coordination problem} arises from the imperfectly complementary of input multi-modality signals, where each image region may involve a different combination of modality (\textit{e.g.}, sky by segmentation masks, tree by sketches+segmentation masks in Figure~\ref{fig:problem} (a)), \textit{i.e.}, the proposed CMCIS task. Such imperfection demands dynamic coordination of modalities (\textit{i.e.}, adaptive fusion) to adapt to varying image regions. The existing methods have limitations in effectively capturing the fine-grained coordination among modalities adapted to different regions (\textit{i.e.}, non-adaptive fusion). They either represent each modality as a single latent vector, thereby sacrificing the regional specificity of modality coordination, or take all modality signals as input then simultaneously learn cross-modal and cross-region interactions, making it challenging to achieve fine-grained adaptation.

And the \textit{modality imbalance problem} arises due to the varying levels of information density exhibited by different modalities,
\textit{e.g.}, text tokens being high density, while segmentation tokens being lower. Such differences over the whole dataset lead to imbalanced distribution of each modality.
During training, the imbalance further affects convergence difficulty~\citep{he2022masked}, which manifests as different modalities converging at different rates and to different endpoints (\textit{e.g.}, sketch converges faster than other modalities in Figure~\ref{fig:problem} (b) because it describes more detailed conditional information and is therefore easier to optimize).
However, existing methods treat each modality equally thus suffering from the issue of modality imbalance, particularly in the proposed CMCIS task.

To tackle the former modality coordination problem, we propose the Mixture-of-Modality-Tokens Transformer (MMoT) to fully exploit the cooperativity across modalities.
Specifically, MMoT uses multiple encoders to model the intra-modal interaction.
Then, modality-specific cross-attention is adopted to inject multimodal conditional information into the decoder.
Finally, the key module multistage token-mixer adaptively fuses multimodal conditioning information with the masked cross-attention mechanism.
To tackle the modality imbalance problem, we propose a multimodal balanced loss to adaptively control the optimization of each modality during the training phase, as well as a multimodal sampling guidance during the sampling phase to control the influences of different modalities and introduce divergence maps to the sampling process to realize more spatially coordinated generation.

To the best of our knowledge, we are the first to focus on specific challenges of the CMCIS task, and our contributions are summarized as follows:
\begin{itemize}
    \item We propose a new challenging task, namely Composed Multimodal Conditional Image Synthesis, which allows users to input various control signals that are not perfectly complementary.
    \item We propose the Mixture-of-Modality-Tokens Transformer (MMoT) for CMCIS, which adaptively fuses fine-grained conditional signals across different modalities.
    \item We introduce the multimodal balanced training loss and the divergence-driven sampling guidance to alleviate the imbalance problem between multiple modalities in CMCIS.
    \item The proposed MMoT accomplishes high-quality image synthesis conditioned on complex compositions of multiple modalities and achieves new state-of-the-art performance on COCO-stuff~\citep{lin2014microsoft, caesar2018coco} and LHQ~\citep{skorokhodov2021aligning}.
\end{itemize}

\section{Related Work}
\subsection{Unimodal Conditional Image Synthesis}
Deep generative models are a family of techniques in which deep neural networks are trained to simulate the distribution of training data.~\citep{bond2021deep}. There are a variety of generative models have been proposed, such as energy-based models~\citep{lecun2006tutorial}, normalizing flows~\citep{papamakarios2021normalizing, kobyzev2020normalizing}, variational autoencoders (VAEs)~\citep{kingma2013auto, sohn2015learning}, generative adversarial networks (GANs)~\citep{goodfellow2020generative, mirza2014conditional}, generative image transformers (GITs)\citep{esser2021taming, chang2022maskgit} and denoising diffusion models~\citep{ho2020denoising, dhariwal2021diffusion}. Generative models typically make trade-offs in quality, sampling speed, and diversity.

GIT is one of the more popular models of late, especially for uni-modal conditional generation~\citep{gafni2022make, yu2022scaling} as the advance in discretizing multi-modalities and the powerful sequence modeling capabilities.
Recently, large-scale text-to-image generative models~\citep{ramesh2022hierarchical, saharia2022photorealistic, yu2022scaling} have made explosive processes and achieved unprecedented superior results.

Traditional GITs~\citep{parmar2018image, child2019generating, chen2020generative} treat image synthesis as a “pixel-by-pixel” autoregressive sequence generation task with the help of the self-attention mechanism~\citep{vaswani2017attention}.
However, as the computation requirement is highly correlated with the sequence length, sampling a high-resolution image may be a challenging endeavor.
The proposed Vector Quantised model~\citep{van2017neural} significantly reduces the processing burden and enables the sampling of high-resolution images based on GITs~\citep{ramesh2021zero, esser2021taming}. And as the scale of the model increases, so does the ability to generate the model~\citep{yu2022scaling}.

The design of model architecture is another point of interest. Instead of using decoder-only language models, recent research~\citep{wu2022nuwa, yu2022scaling} employs an encoder-decoder transformer for conditional image synthesis and achieves promising results. 
Our work proposed a novel encoder-decoder-based architecture that decouples intra-modal interaction and fusion.

\subsection{Multimodal Conditional Image Synthesis}
Multimodal conditional image synthesis has attracted increasing attention recently. Representative work includes M6-UFC~\citep{zhang2021ufc} and PoE-GAN~\citep{huang2022multimodal}.

M6-UFC is a Bert-based framework based on the two-stage image synthesis method~\citep{van2017neural, razavi2019generating, chen2020generative, esser2021taming, ramesh2021zero}. In M6-UFC, the multimodal conditional inputs and generated image are transformed into a sequence of tokens to be processed by the unidirectional Transformer decoder~\citep{vaswani2017attention}. The advantages of M6-UFC are that it unifies various modalities in a universal form and can thus easily extend to more guidance modalities. However, it employs concatenation to combine multimodal user inputs, and intra-modal interaction is interlaced with fusion; as a result, it may struggle with handling missing modalities~\citep{ma2022multimodal, huang2022multimodal}.

PoE-GAN is GANs based method with the multiscale projection discriminator~\citep{miyato2018cgans, liu2019learning, wang2018high}. In PoE-GAN, conditional information from multiple modalities is first encoded into a unified latent space and then fused using product-of-experts modeling~\citep{hinton2002training}. The advantages of PoE-GAN are that it decouples intra-modal interaction and fusion, and is more robust to missing modalities. However, conditional GANs are known to be susceptible to mode collapse~\citep{isola2017image, odena2017conditional} and spatial information is lost in the latent space of PoE-GAN.
MMoT inherits the advantages of the above two works and is able to produce reasonable outputs even with counterfactual condition modalities.

Recently, there have been some methods~\citep{huang2023composer, zhang2023adding, mou2023t2i} that attempt to introduce more control signals to large-scale Text-to-Image pre-trained models (\textit{e.g.}, Stable diffusion~\citep{rombach2022high}) to achieve multimodal conditional image generation. However, such methods are text-centric, so it is difficult to leverage the impact of various imperfectly complementary modality in the case of our proposed CMIS task.

\section{Method}
\begin{figure*}
    \centering
    \includegraphics[width=\linewidth]{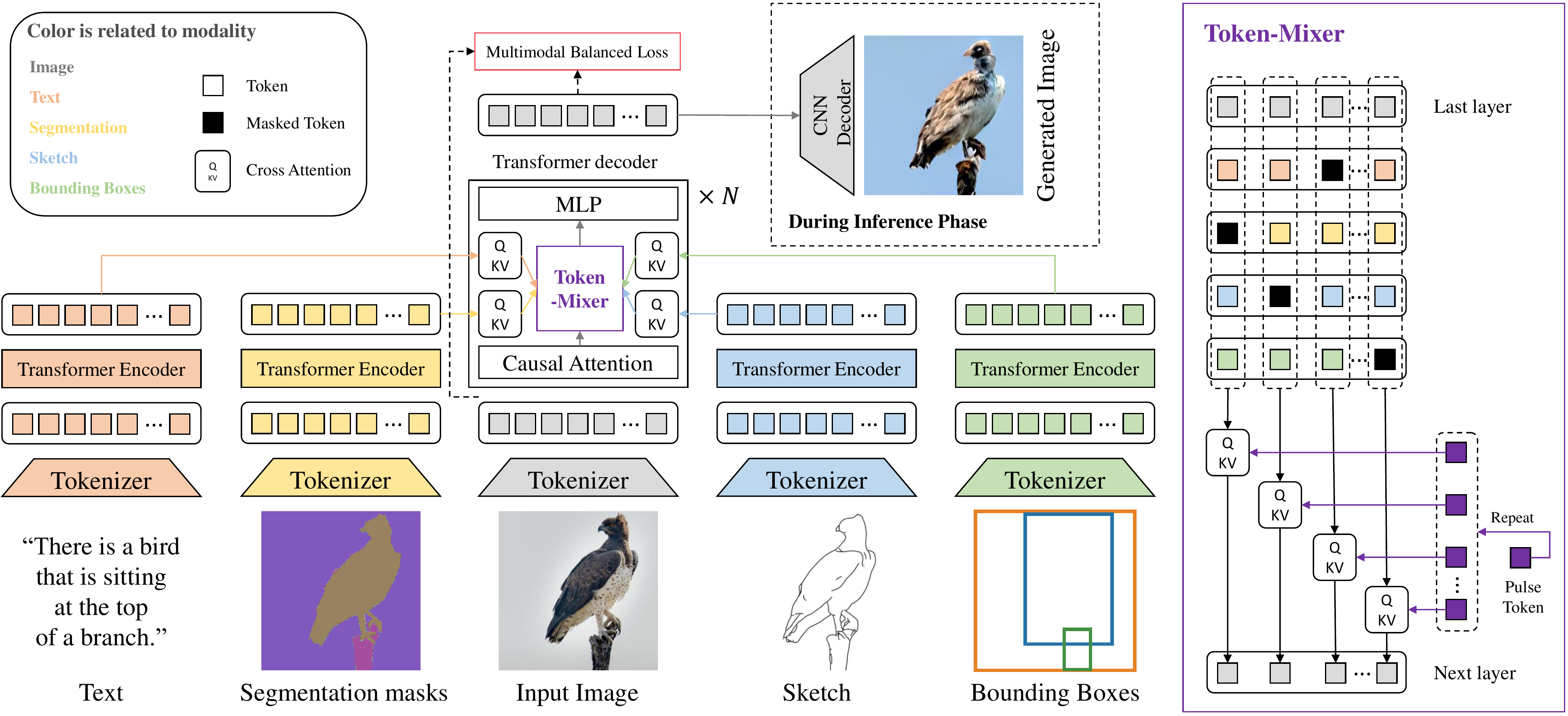}
    \caption{The overview of the Mixture-of-Modality-Tokens Transformer for CMCIS task. Given an image and multiple modalities, including text, segmentation masks, sketch, and bounding boxes, we tokenize them into discrete tokens with different tokenizers, and then (a) model intra-modal interaction with modality-specific encoders; (b) inject multimodal conditioning information into the decoder with modality-specific cross-attention; (c) adaptively fuse conditional signals via the multistage token-mixer. We train MMoT with multimodal balanced loss and sample with divergence-driven multimodal guidance.}
    \label{fig:framework}
\end{figure*}

The goal of CMCIS is to train a single model that approximates image distributions conditional on any combination of feasible modalities. Mathematically, the objective is to learn $p(x\vert\mathcal{C})$ when given a dataset of images $x$ paired with a conditional input $\mathcal{C}$ which consists of $M$ different modalities $\mathcal{C}\subseteq \{c_1, c_2, \dots, c_M\}$.

Our model follows an encoder-decoder structure, consisting of $M$ encoders for modeling intra-modal interaction and a common decoder for fusion among multi-modalities, as shown in Figure~\ref{fig:framework}. Our proposed methods are described in detail later, comprised of 1) describing the two-stage transformer-based framework, 2) introducing the Mixture-of-Modality-Token transformer with a multistage token-mixer module, 3) training with multimodal balanced loss, and 4) sampling with divergence-driven guidance. 

\subsection{Preliminary}
In this section, we review the two-stage transformer-based framework~\citep{van2017neural, razavi2019generating, chen2020generative, esser2021taming, ramesh2021zero} for image synthesis.
At the first stage, a convolutional model consisting of an encoder $E$ and a decoder $D$ is learned, such that, an image $x \in \mathbb{R}^{3 \times H \times W}$ can be represented as a collection of code-words $z_\mathbf{q} \in \mathbb{R}^{n_z \times h \times w}$ with code from a learned, discrete codebook $\mathcal{Z} = \{z_k \}_{k=1}^{K}$, where $n_z$ is the dimensionality of codes, $z_k \in \mathbb{R}^{n_z}$ is the $k^{th}$ code-word, and $K$ is the number of code-words. 
More precisely, the convolutional encoder $E$ first encodes the $x$ as $\hat{z}=E(x) \in \mathbb{R}^{n_z \times h \times w}$ and then an element-wise quantization $\mathbf{q}(\cdot)$ is applied to each spatial element $\hat{z}_{ij} \in \mathbb{R}^{n_z}$ to obtain the closest discrete code-work $z_k$, \textit{i.e.}, $\mathbf{q}(\hat{z}_{ij})=\text{arg\,min}_{z_k \in \mathcal{Z}} \Vert \hat{z}_{ij} - z_k \Vert$. At last, a given image $x$ can be approximated by a convolutional decoder $D$, \textit{i.e.}, $\hat{x}=D(z_{\mathbf{q}})$. Overall, the first stage can be denoted by:
\begin{equation}
    z_\mathbf{q}=\mathbf{q}(E(x)), \hat{x}=D(z_{\mathbf{q}}),
\end{equation}
the convolutional model and the codebook can be trained via the loss function:
\begin{align}
\mathcal{L_{VQ}}(E, D, \mathcal{Z}) = \Vert x - \hat{x} \Vert^2 
  &+ \Vert \text{sg}[E(x)] - z_{\mathbf{q}} \Vert_2^2 \nonumber \\ &+ \Vert \text{sg}[z_{\mathbf{q}}] - E(x) \Vert_2^2,
\label{eq:origvqloss}
\end{align}
where $\text{sg}[\cdot]$ denotes the stop-gradient operation.

At the second stage, the quantized encoding of an image $x$ can be transformed into a sequence of code-word index $s \in \{0, \dots, \vert \mathcal{Z} \vert - 1 \}^{h \times w}$. Therefore, image generation can be formulated as autoregressive sequence generation modeled with a transformer. The transformer learns to predict the distribution $p(s_i \vert s_{<i})$ of possible next token $s_i$ given preceding tokens $s_{<i}$. Then the likelihood of the full representation can be computed as $p(s)=\prod\nolimits_i p(s_i \vert s_{<i})$. The transformer can be trained by maximizing the log-likelihood:
\begin{equation}
    \mathcal{L}_{T} = \mathbb{E}_{x \sim p(x)} \left[-\log p(s) \right].
\end{equation}

\subsection{Mixture-of-Modality-Tokens Transformer}
We propose a novel framework, namely Mixture-of-Modality-Tokens Transformer (MMoT) for adaptive fusion among multimodal information.

\subsubsection{Modality representation in unified-form}
In this paper, we consider four commonly used input conditions, including text, segmentation mask, sketch, and bounding box layout.
For visual modalities, \textit{i.e.}, image, segmentation masks, and sketch, we quantize it with the help of discrete VAE, \textit{e.g.}, VQGAN~\citep{esser2021taming}. 
For bounding boxes layout, we follow the same tokenization method as in~\citep{jahn2021high}.
And CLIP~\citep{radford2021learning} is our solution for tokenizing text modalities.

Specifically, to unify the embeddings of different modalities including image, segmentation masks, sketch, bounding boxes, and text, we treat all of them as language-like tokens and embed the tokens to the latent space $U_m \in \mathbb{R}^{l \times d}$, where $m$ is used to distinguish different modalities, $l$ denotes the number of tokens and $d$ means the dimension of each embedding. 

Given an image $\mathcal{X} \in \mathbb{R}^{3 \times H_I \times W_I}$, we quantize it with the help of a discrete VAE, e.g., VQGAN~\citep{esser2021taming}.
Specifically, the discrete VAE encodes the image to a latent space, and maps the latent features to the closest discrete tokens from the codebook $\mathcal{Z}_{image} = \{z_k \}_{k=1}^{K}$ with $K$ entries. The quantized representations are in the format of $\mathbb{N}_{image}^{\{h_i, w_i\}}$, where $\mathbb{N}_{image}$ is the set of integers in $[0,1,\dots,K)$, and $h_i, w_i$ are the dimension of quantized representations.
The segmentation masks $\mathcal{S} \in \mathbb{R}^{H_S \times W_S}$ and the sketch $\mathcal{K} \in \mathbb{R}^{3 \times H_K \times W_K}$ can be tokenized in the same pipeline but with different pre-trained discrete VAE models. Such content of segmentation masks and sketch in the raw space can be represented as $s \in \mathbb{N}_{seg}^{\{h_s, w_s\}}$ and $k \in \mathbb{N}_{sketch}^{\{h_k, w_k\}}$ with different codebook $\mathbb{N}_{seg} \in [0,1,\dots, K')$ and $\mathbb{N}_{sketch} \in [0,1,\dots, K^{\dag})$, respectively.

Given a bounding boxes layout consists of a set of objects of their positions and category classes, we directly tokenize it into sequential object tokens $b=\{(o_i, p_i)_{i=1}^{N_B} \}$ with $N_B$ objects, where $o_i$ denotes the $i^{th}$ object’s category, and $p_i = [tl_i, br_i] $ represents its top-left and bottom-right corner positions.

For image, segmentation masks, sketch, and bounding boxes, we feed their discrete codes to respective embedding layers to get the continuous representations.

As for text modalities, CLIP~\citep{radford2021learning} is our solution. A transformer-based encoder is used to represent the discrete token sequence into a high dimensional vector once the sentence is tokenized with Byte-Pair Encoding (BPE). Then we use this vector as the final representation in the form of $\mathbb{R}^{\{1,d\}}$, which means the text embedding only occupies one token as the input to our MMoT module.
For datasets that lack corresponding textual descriptions, we use image embedding as the pseudo-text representation from CLIP as a replacement. 

In general, the representations of images can be denoted as $X \in \mathbb{R}^{l_{image} \times d}$ and the representations of the conditional modality $m$ can be denoted in the same dimension as $C_m \in \mathbb{R}^{l_m \times d}$ but with different length of tokens $l_m$, and as we discussed, the text modality contributes only one token, where $l_{text}=1$.

\subsubsection{Attention mechanism} 
We recall the attention mechanism since it is an important means for MMoT to achieve interaction and fusion. The attention mechanism draws global dependencies between them with Query-Key-Value (QKV) model, where queries $Q=W_QX$, keys $K=W_KC$ and values $V=W_VC$ with the learnable weights $W$. Thus the attention function can be defined as:
\begin{equation}
    \text{Attn}(X, C) = softmax(\frac{QK^T}{\sqrt{D}})V.
\end{equation}

It is worth noting that with the given condition $C$ that differs from the input $X$, the attention mechanism is widely known as cross-attention $\text{CA}(X, C)$. If the condition information $C$ is the same as the input $X$, the attention mechanism can be expressed as $\text{Attn}(X, X)$, which is known as self-attention $\text{SA}(X)$.

\subsubsection{Modeling intra-modal interaction with modality-specific encoders}
To model the intra-modal interaction for various modalities, we introduce modality-specific encoders that project input features into an intermediate representation.

Each modality encoder consists of $N$ self-attention layers. Given the input features $C_m^{e}(n-1) \in \mathbb{R}^{L_m \times D}$ of modality $m$ in $(n-1)^{th}$ layer, the output $C_m^{e}(n) \in \mathbb{R}^{L_m \times D}$ of the $n^{th}$ encoder layer can be calculated as:
\begin{equation}
    C_m^{e}(n) = \text{SA}[C_m^{e}(n-1)].
\end{equation}

\subsubsection{Injecting multimodal conditioning information with modality-specific cross attention}
In order to inject multimodal conditional information into the decoder, we use modality-specific cross-attention to fuse the image feature with each modality feature.

Specifically, given the input image features $X(n-1)$ of the $(n-1)^{th}$ decoder layer and the output $C_m^{e}(N)$ of the encoder's last layer, the output $C_m^{d}(n) \in \mathbb{R}^{L_{image} \times D}$ for $n^{th}$ decoder layer is given by:
\begin{equation}
\begin{aligned}
X(n) &= \text{SA}[X(n-1)], \\
       C_m^{d}(n) &= \text{CA}[X(n), C_m^{e}(N)].
\end{aligned}
\end{equation}

\subsubsection{Adaptive fusion with multistage token-mixer}
After applying cross-attention, the multistage token-mixer is proposed to fuse the modality tokens which contain conditional information related to a specific modality. A special \texttt{[PULSE]} token $P$ within token-mixer is introduced to adaptively estimate the combination weights (\textit{i.e.}, attention scores) of each modality token and fuse them with the masked cross-attention mechanism. The combination-weight maps in Figure~\ref{fig:analysis} (c) show that \texttt{[PULSE]} token can effectively evaluate the influences of different modalities in different decoder layers.

Specifically, with the output $X(n)$ of the $n^{th}$ self-attention layer and a set of outputs of the $n^{th}$ cross-attention layer, we adapt a multistage token-mixer to fuse the conditional information from different modalities and then feed the fusion features to the subsequent decoder layers:
\begin{equation}
\begin{aligned}
    X^{\dag}(n) = \text{Mixer}[X(n), \bm{C}^{d}(n)],
\end{aligned}
\end{equation}
where $\bm{C}$ is the stack of all conditional modalities, \textit{i.e.}, $\bm{C} \in \mathbb{R}^{M \times L_{image} \times D}$.

The Mixer function can be defined as:
\begin{equation}
    \text{Mixer}(X, \bm{C}) := softmax(\frac{PF^T}{\sqrt{D}})F,
    \label{eq:mixer_function}
\end{equation}
where we concatenate the latent representation of image $X(n)$ with the stack representation $\bm{C}$ of $M$ modalities along the modality dimension to form $F \in \mathbb{R}^{L_{image} \times (M+1) \times D}$, while its transpose can be denoted as $F^T \in \mathbb{R}^{L_{image} \times D \times (M+1)}$. $P \in \mathbb{R}^{L_{image} \times 1 \times D}$ is the \texttt{[PULSE]} token. Noted that the random masks will be applied in the Mixer function during the training phase, which serve as modality dropouts to handle
missing-modality cases during inference.

\subsection{Multimodal Balanced Loss}
To train the MMoT, we propose $\mathcal{L}_{mmb}$, an improved cross-entropy loss named multimodal balanced loss, for sequential prediction tasks to realize balanced optimization among different modalities:
\begin{equation}
    \mathcal{L}_{mmb} = \mathbb{E}_{s \sim p(s)} \mathbb{E}_{x, \mathcal{C}_s \sim p(x, \mathcal{C}_s)} \left[-\log p(x \vert \mathcal{C}_s) \right],
    \label{eq:loss}
\end{equation}
where $p(s)$ is the probability of the occurrence of subset $\mathcal{C}_s$. In order to adaptively control the optimization of each subset, we set:
\begin{equation}
    p(s) = -\log p(x \vert \mathcal{C}_s) \bigg/ \sum\nolimits_{k=1}^{2^M} \left[-\log p(x \vert \mathcal{C}_k) \right].
\end{equation}
As the joint conditional distribution in Eq.~(\ref{eq:loss}) is able to be represented as the product of sequential conditional distributions in an auto-regressive process, we can have:
\begin{equation}
    p(x \vert \mathcal{C}_s) = \prod_i p(x_i \vert x_{<i}, \mathcal{C}_s),
    \label{eq:ar}
\end{equation}
where $i$ is the index in a token sequence. In the simplest case, $p(s)=1/2^M$, which means that any subset $\mathcal{C}_s$ of input modalities has the same chance to appear in the forward process, would cause an imbalanced multimodal optimization because different subsets contain different levels of control information. 
It is therefore necessary to introduce a parameter to indicate the strength of the given conditions.
Intuitively, $p(x \vert \mathcal{C}_s)$ indicates how easy it is to optimize this subset, so that when $p(s)$ proportional to $\left[-\log p(x \vert \mathcal{C}_s) \right]$ can make MMoT focus on the subsets that are more difficult to optimize. We will show the effectiveness of multimodal balanced loss in ablation experiments.

\subsection{Divergence-driven Sampling Guidance}
During inference, we propose multimodal guidance for the CMCIS task to balance the influences of various control signals. Assuming that all the input conditional modalities are statistically independent and using $M'$ to denote the number of modalities in subset $\mathcal{C}_s$, the sequential conditional distribution $p(x_i \vert x_{<i}, \mathcal{C}_s)$ in Eq.~(\ref{eq:ar}) can be rewritten as follows:
\begin{equation}
    \begin{aligned}
        p(x_i \vert x_{<i}, \mathcal{C}_s) &= p(x_i \vert x_{<i}) \left[\frac{p(\mathcal{C}_s \vert x_i, x_{<i})}{p(\mathcal{C}_s \vert x_{<i})} \right] \\
        &= p(x_i \vert x_{<i}) \prod_{m=1}^{M'} \left[\frac{p(c_m \vert x_i, x_{<i})}{p(c_m \vert x_{<i})} \right].
    \end{aligned}
\end{equation}

Inspired by the influence of the conditioning signal can be amplified by the guidance scale~\citep{dhariwal2021diffusion}, we use $\lambda_m$ to control the influence of the $m^{th}$ conditioning signals:
\begin{equation}
        \begin{aligned}
        p_\lambda(x_i \vert x_{<i}, \mathcal{C}_s) &\propto p(x_i \vert x_{<i}) \prod_{m=1}^{M'} \left[\frac{p(c_m \vert x_i, x_{<i})}{p(c_m \vert x_{<i})} \right]^{\lambda_m} \\
        &= p(x_i \vert x_{<i}) \prod_{m=1}^{M'} \left[\frac{p(x_i \vert x_{<i}, c_m)}{p(x_i \vert x_{<i}, \emptyset)} \right]^{\lambda_m}.
    \end{aligned}
    \label{eq:arclg}
\end{equation}

For mitigation of computation, we denote Eq.~(\ref{eq:arclg}) to the log in logarithmic form:
\begin{equation}
    \begin{aligned}
        \log p_\lambda&(x_i \vert x_{<i}, \mathcal{C}_s) = \log p(x_i \vert x_{<i}) + \\
        &\sum_{m=1}^{M'} \lambda_m \left[ \log p(x_i \vert x_{<i}, c_m) - \log p(x_i \vert x_{<i}, \emptyset) \right],
    \end{aligned}
    \nonumber
\end{equation}
we can then synchronously generate multiple parallel token streams: token streams conditioned on different modalities including empty input, and apply multimodal guidance on logit scores:
\begin{equation}
\begin{aligned}
    &\text{p}^{uncon} = \text{TL} (x \vert \emptyset), \\
    &\text{p}^{con}_m = \text{TL} (x \vert c_m), \\
    \text{p} = &\text{p}^{uncon} + \sum_{m=1}^{M'} \lambda_m (\text{p}^{con}_m - \text{p}^{uncon}),
\end{aligned}
\end{equation}
where the function $\text{TL}(x\vert c)$ computes the logits outputted by MMoT decoder when conditioned on $c$, $\emptyset$ means the null condition for classifier free, $\text{p}^{uncon}$ are logits scores obtained by unconditional token stream, $\text{p}^{con}_m$ are logits scores obtained by conditional token stream of modality $m$, $\text{p}$ are the multimodal guided logits score, and $\lambda_m$ is the guidance scale relevant to the corresponding modality.

In addition, based on an observation that the Jensen–Shannon Divergence (JSD) between the unconditional logits and conditional logits contains rich semantic information (Figure~\ref{fig:analysis} (b)), we use JS divergence to decide the multimodal guidance scale:
\begin{equation*}
    \lambda_m \propto \text{JSD} (\text{p}^{con}_m - \text{p}^{uncon}).
\end{equation*}

It is worth noting that the suggested multimodal guidance can not only increase sample quality but, more crucially, lead to more spatially coordinated images.

\section{Experiments}
In this section, we evaluate the quality and diversity of our versatile MMoT in synthesizing images under various conditional modalities or compositions of them.  The generated images with a set of conditional input modalities show that MMoT can carry out effective interaction and fusion of multimodal information (Sect.~\ref{sec:multimodal samples}), and superior results with extensive conditional image synthesis methods state that MMoT is robust to all kinds of modalities (Sect.~\ref{sec: comparisons}). We conduct ablation studies to validate the effectiveness of different modules of MMoT (Sect.~\ref{sec: ablations}), and we provide some important insights about how MMoT realizes interaction and fusion via several visualizations (Sect.~\ref{sec: analysis}).

We performed our experiments on two datasets: COCO-Stuff~\citep{lin2014microsoft, caesar2018coco} and LHQ~\citep{skorokhodov2021aligning}. COCO-Stuff is a derivative work of the COCO dataset, which contains dense pixel-level and instance-level annotations including text descriptions, segmentation maps, bounding boxes, and so on. LHQ is a dataset of 90k nature landscapes but without any annotations, so we use pseudo-labeling methods to obtain text, segmentation masks, and sketch annotations. More details about the datasets are in~\ref{sec:details-datasets}.

\subsection{Experimental Setup}
\subsubsection{Datasets}\label{sec:details-datasets}
We evaluate the proposed MMoT on COCO-Stuff~\citep{lin2014microsoft, caesar2018coco} and LHQ~\citep{skorokhodov2021aligning}. All input modalities are obtained from either human annotations or pseudo-labeling methods. 
And for fair comparisons with PoE-GAN, the same pseudo-labeling methods were used in our approach.
We describe details about each dataset in the following.

\noindent\textbf{COCO-Stuff}
is an expansion of the Microsoft Common Objects in Context (MSCOCO) dataset~\citep{lin2014microsoft}, which includes 91 stuff classes and 80 object classes. It contains 123,287 images of complex scenes, including 118,287 training images and 5,000 test images. All images are randomly cropped to 256$\times$256 in our main experiments and ablation studies.

Annotations (\textit{i.e.}, text, segmentation mask, sketch, and bounding box layout) for each image are obtained from either human annotations or pseudo-labeling methods: \textbf{(i)} In COCO-Stuff, each image has 5 \textit{text} captions, we use CLIP text encoder to extract a high dimension vector per caption. \textbf{(ii)} We direct use the \textit{segmentation mask} provided in COCO-Stuff. \textbf{(iii)} To obtain the \textit{sketch} annotation, we first detect the edge per image with the HED~\citep{xie2015holistically} and then simplify the rough sketch with the sketch cleanup process~\citep{simo2016learning}. \textbf{(iv)} We use the bounding boxes and labels provided in COCO-stuff as the ground truth \textit{bounding box layout}.

\noindent\textbf{LHQ} is a dataset containing 90,000 high-resolution landscape images crawled and preprocessed from Unsplash and Flickr. The dataset is randomly split into an 86,400 training set and a 3,600 test set, and all images are randomly cropped to 256$\times$256 in our main experiments. 

Since the vanilla dataset does not come with any manual annotations, annotations (\textit{i.e.}, text, segmentation mask, and sketch) are obtained from pseudo-labeling methods: \textbf{(i)} For the \textit{text} annotation, we use the pre-trained CLIP image encoder to extract a feature vector as the pseudo text embedding. \textbf{(ii)} DeepLab-v2~\citep{chen2017deeplab} was used to produce pseudo \textit{segmentation mask} annotation. \textbf{(iii)} HED~\citep{xie2015holistically} followed by the sketch cleanup process~\citep{simo2016learning} was adopted to annotate each image with a \textit{sketch} map.

\subsubsection{Evaluation metrics}
For different conditional image synthesis tasks, we use different metrics to evaluate the generation performance over all existing methods and our proposed MMoT. They are Inception Score (IS)\citep{salimans2016improved}, Frechet Inception Distance (FID)\citep{heusel2017gans} and Clean-FID\citep{parmar2021buggy}. IS and FID are the most commonly used metrics to evaluate the quality and diversity of generated images. Clean-FID is an improved version of FID.

\noindent\textbf{Inception Score (IS)} measures the quality of generated images by computing the expected KL-divergence between the marginal class distribution over all generated images and the conditional distribution for a particular generated image, using the class probability predicted by the Inception Net. This metric is expected to capture both the fidelity and diversity of generated images.

\noindent\textbf{Frechet Inception Distance (FID)} measures the similarity between the embedding feature of generated and real images. This is achieved by fitting the embedding features into a multivariate Gaussian distribution and computing their Frechet distance.

\noindent\textbf{Clean-FID.} Under the hood, computing FID contains several subtle implementation decisions, notably image resizing, quantization, and formatting. Any inconsistency in the steps leads to results that are no longer comparable to other methods. The resize operation and the image quantization/compression are especially impactful. To facilitate an easy comparison, \citep{parmar2021buggy} propose an easy-to-use library, \textit{i.e.}, clean-fid, which is more suitable for benchmarking due to its reported benefits over previous implementations of FID.

\subsubsection{Hyper-parameters}
The number of encoder layers in the MMoT is 12, while the number of decoder layers is 24. We use this asymmetrical structure because the encoder is mainly used to extract features, while the decoder is responsible for more complex mapping, \textit{i.e.}, converting the input features into the final image output, and this design can greatly reduce the number of parameters.
Encoders for different modalities have the same number of layers.
Both encoders and decoders share an architecture of 12 attention heads and an embedding dimension of 768.
All training images in COCO-Stuff and LHQ are randomly cropped to $256 \times 256$ in our main experiments and ablation studies.
We use the AdamW~\citep{loshchilov2017decoupled} optimizer with $\beta_1 = 0.9, \beta_2 = 0.95$ and the weight decay is set to be 0.01.
We use a batch size of 64 for training all our models and set the learning rate to be 4.5e-6, and all the models are trained for 300 epochs on 8 A100 GPUs.

\subsection{Comparisons with Existing Methods}\label{sec: comparisons}
\begin{table}[t]
\begin{center}
\begin{minipage}{\linewidth}
\caption{Comparison on COCO-Stuff (256$\times$256)}\label{tab:main_coco}
\resizebox{\textwidth}{15mm}{
    \begin{tabular}{lccclcc}
    \toprule
    \multicolumn{3}{c}{(a) Text to Image} & & \multicolumn{3}{c}{(b) Bounding boxes to Image} \\
    \cmidrule{1-3} \cmidrule{5-7}
    \textbf{Method} & FID $\downarrow$ & Clean-FID $\downarrow$ & & \textbf{Method} & FID $\downarrow$ & Inception Score $\uparrow$ \\
    \cmidrule{1-3} \cmidrule{5-7}
    - & - & - &                           & LostGAN-V2 & 42.6 &18.0±0.50                            \\
      -  & - & - &                          & OC-GAN & 41.7 & 17.8±0.00                             \\
    DF-GAN  & - & 45.2 &                  & VQGAN+T & 33.7 & -                                      \\
      DM-GAN+CL & - & 29.9 &                & LAMA & 31.1 & -                                       \\
    VQGAN+T* & \underline{27.8} & 28.1 &  & Context-L2I & 29.56 & 18.57±0.54                        \\
    PoE-GAN & - & \underline{20.5} &      & TwFA & \underline{22.1} & \underline{24.3±1.04}         \\
    \midrule
    \emph{MMoT(Ours)} & \textbf{17.9} & \textbf{17.8} & & \emph{MMoT} & \textbf{19.2} & \textbf{26.7±0.50} \\
    \botrule
    \end{tabular}}
\resizebox{\textwidth}{15mm}{
    \begin{tabular}{lccclcccc}
    \toprule
    \multicolumn{3}{c}{(c) Segmentation masks to Image} & & \multicolumn{3}{c}{(d) Sketch to Image} & & (e) All$^\dagger$ \\
    \cmidrule{1-3} \cmidrule{5-7} \cmidrule{9-9}
    \textbf{Method} & FID $\downarrow$ & Clean-FID $\downarrow$ & & \textbf{Method} & FID $\downarrow$ & Clean-FID $\downarrow$ & & Clean-FID $\downarrow$ \\
    \cmidrule{1-3} \cmidrule{5-7} \cmidrule{9-9}
    pix2pixHD & 111.5 & - &               & - & - & - & & - \\
    SPADE & 22.6 & 22.1 &                 & pix2pixHD* & 44.4 & 46.2 & & - \\
    VQGAN+T & 22.4 & 21.6 &               & SPADE* & 78.8 & 80.3 & & - \\    
    OASIS & 17.0 & 19.2 &                 & VQGAN+T* & 33.9 & 34.4 & & - \\     
    PITI & \underline{15.8} & - &         & PITI & \textbf{20.3} & - & & - \\ 
    PoE-GAN & - & \underline{15.8} &      & PoE-GAN & - & \underline{25.5} & & \underline{13.6} \\        
    \midrule
    \emph{MMoT} & \textbf{12.7} & \textbf{12.9} & & \emph{MMoT} & \underline{23.1} & \textbf{23.9} & & \textbf{12.6} \\
    \botrule
    \end{tabular}}
\footnotetext{We evaluate models conditioned on different modalities (\textit{i.e.}, text, bounding boxes, segmentation masks, sketch). The best scores are highlighted in bold and the second best ones are underlined. For fair comparisons, all the results are taken from the relative papers. `-' means the related value is unavailable in their papers. `*' denotes results on samples from retrained models with the official implementation. All$^\dagger$ means image synthesis conditioned on text+segmentation masks+sketch.}
\end{minipage}
\end{center}
\end{table}

\begin{table}[t]
\begin{center}
\begin{minipage}{\linewidth}
\caption{Comparison on LHQ (256$\times$256)}\label{tab:main_lhq}
\resizebox{\textwidth}{14mm}{
    \begin{tabular}{lccc}
    \toprule
    \multicolumn{4}{c}{(a) Text to Image} \\
    \cmidrule{1-4}
    \textbf{Method} & FID $\downarrow$ & Clean-FID $\downarrow$ & Inception Score $\uparrow$ \\
    \cmidrule{1-4}
    VQGAN+T & 12.71 & 12.78 & 4.61                          \\
    MaskGIT\dag & 24.33 & - & 4.61                          \\
    NUWA-Infinity\dag & \textbf{9.71} & - & \textbf{4.98}   \\
    \midrule
    \emph{MMoT(Ours)} & \underline{11.38} & \textbf{11.46} &\underline{4.94 ± 0.19} \\
    \bottomrule
    \end{tabular}}
\resizebox{\textwidth}{14mm}{
    \begin{tabular}{lccc}
    \toprule
    \multicolumn{4}{c}{(b) Segmentation masks to Image} \\
    \cmidrule{1-4}
    \textbf{Method} & FID $\downarrow$ & Clean-FID $\downarrow$ & Inception Score $\uparrow$ \\
    \cmidrule{1-4}
    pix2pixHD & 36.52 & 41.00 & 3.42±0.10      \\
    SPADE & 25.47 & 26.70 & 3.71±0.12          \\
    VQGAN+T & 14.92 & 15.00 & 4.42±0.11        \\
    \midrule
    \emph{MMoT} & \textbf{11.87} & \textbf{11.98} &\textbf{4.43 ± 0.13} \\
    \bottomrule
    \end{tabular}}
\resizebox{\textwidth}{14mm}{
    \begin{tabular}{lccc}
    \toprule
    \multicolumn{4}{c}{(c) Sketch to Image} \\
    \cmidrule{1-4}
    \textbf{Method} & FID $\downarrow$ & Clean-FID $\downarrow$ & Inception Score $\uparrow$ \\
    \cmidrule{1-4}
    pix2pixHD & 32.19 & 36.23 & 3.51±0.13 \\
    SPADE & 35.83 & 36.35 & 2.98±0.07  \\
    VQGAN+T & 13.91 & 13.9 & 4.13 ± 0.21 \\
    \midrule
    \emph{MMoT} & \textbf{11.68} & \textbf{11.76} &\textbf{4.26 ± 0.09} \\
    \bottomrule
    \end{tabular}}
\footnotetext{We evaluate models conditioned on different modalities (\textit{i.e.}, text, segmentation masks, sketch). The best scores are highlighted in bold and the second best ones are underlined. All the results are on samples from retrained models with the official implementation. `\dag' denotes results taken from the relative papers which trained on LHQC with 1024$\times$1024 resolution. `-' means the related value is unavailable in their papers.}
\end{minipage}
\end{center}
\end{table}

\begin{figure*}
    \centering
    \includegraphics[width=\textwidth]{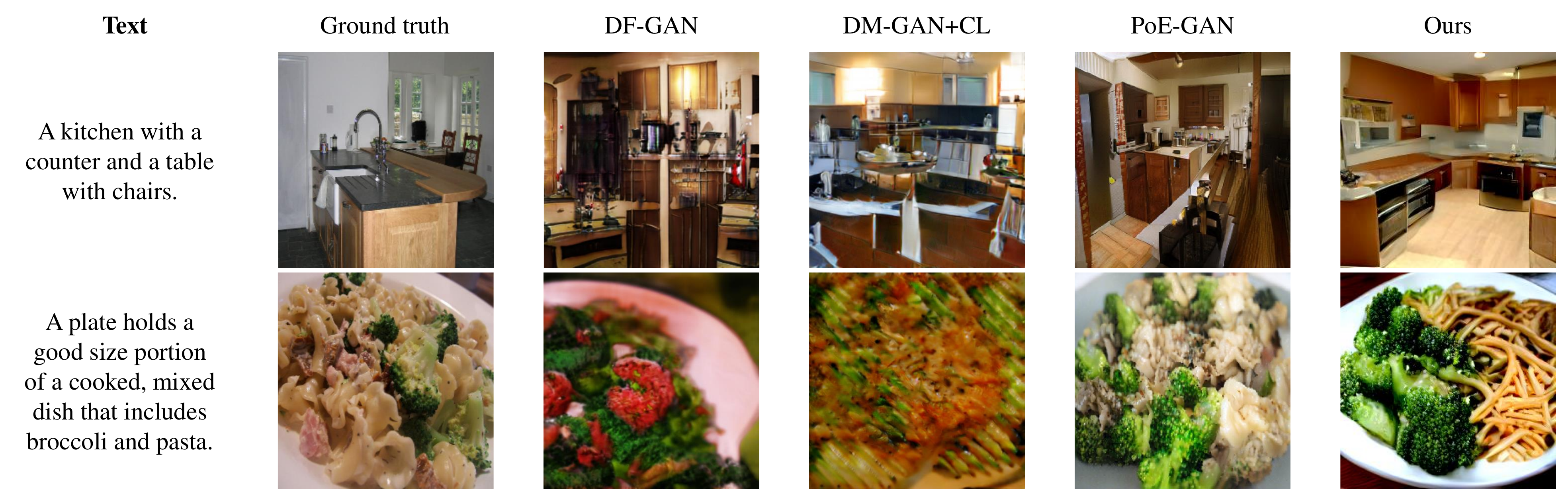}
    \caption{Qualitative comparison of text-to-image synthesis results on COCO-Stuff. More results are demonstrated in~\ref{sec:qualitative-ucis}}
    \label{fig:text-to-image}
\end{figure*}
\begin{figure*}
    \centering
    \includegraphics[width=\textwidth]{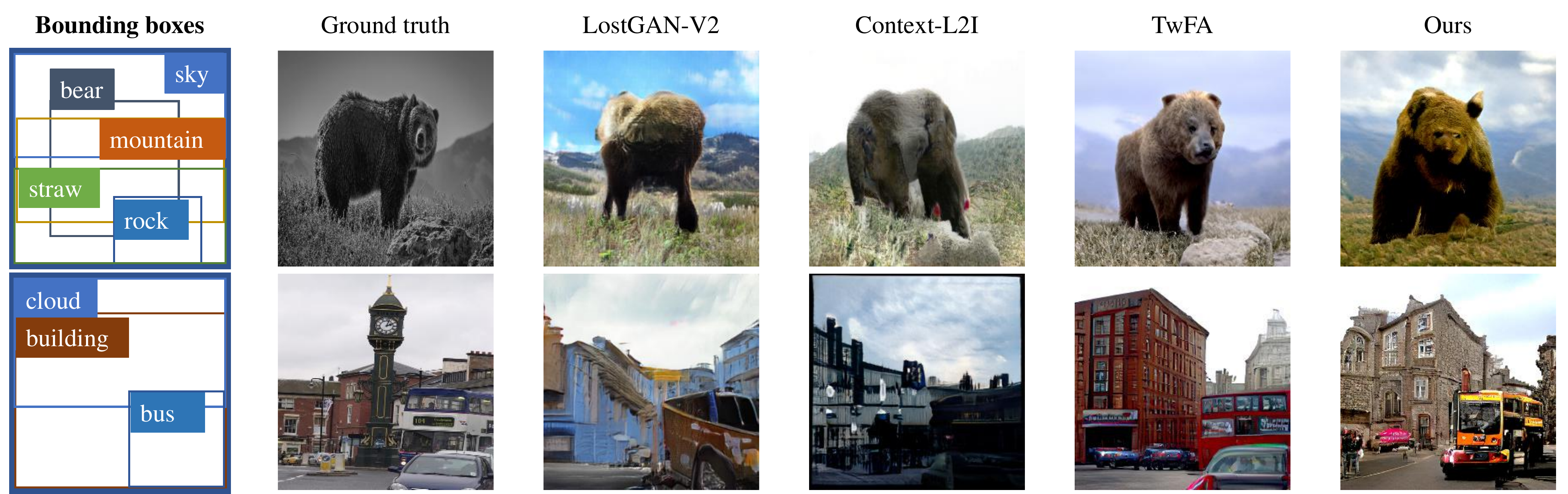}
    \caption{Qualitative comparison of bounding boxes-to-image synthesis results on COCO-Stuff. More results are demonstrated in~\ref{sec:qualitative-ucis}}
    \label{fig:layout-to-image}
\end{figure*}
\begin{figure*}
    \centering
    \includegraphics[width=\textwidth]{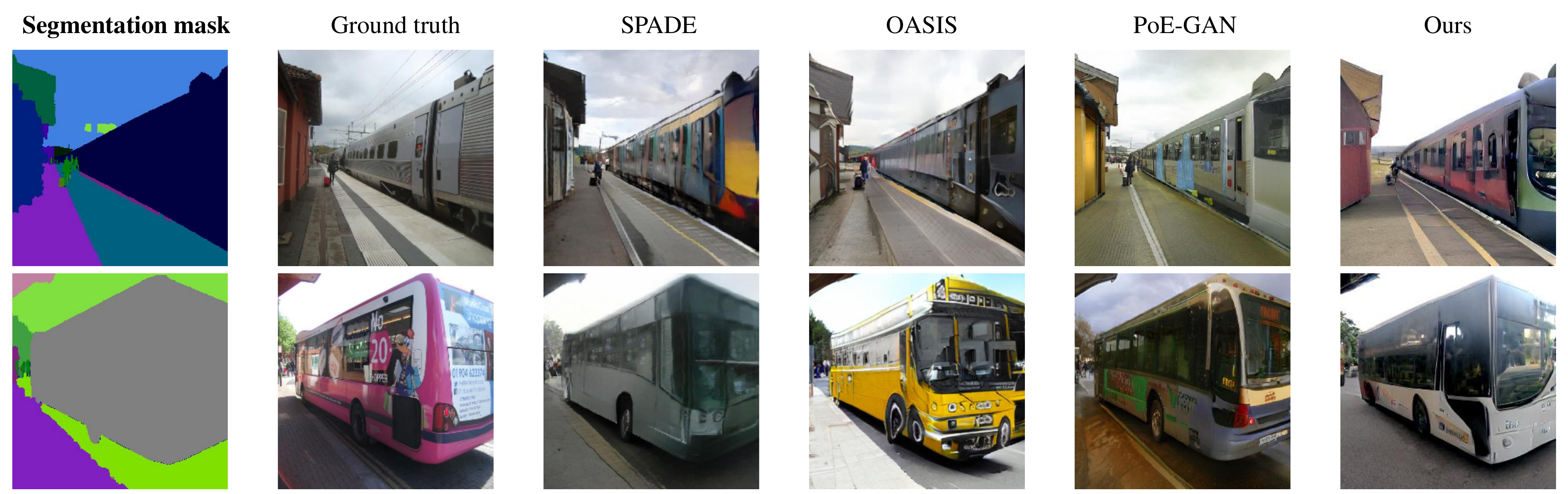}
    \caption{Qualitative comparison of segmentation-to-image synthesis results on COCO-Stuff. More results are demonstrated in~\ref{sec:qualitative-ucis}}
    \label{fig:seg-to-image}
\end{figure*}
\begin{figure*}
    \centering
    \includegraphics[width=\textwidth]{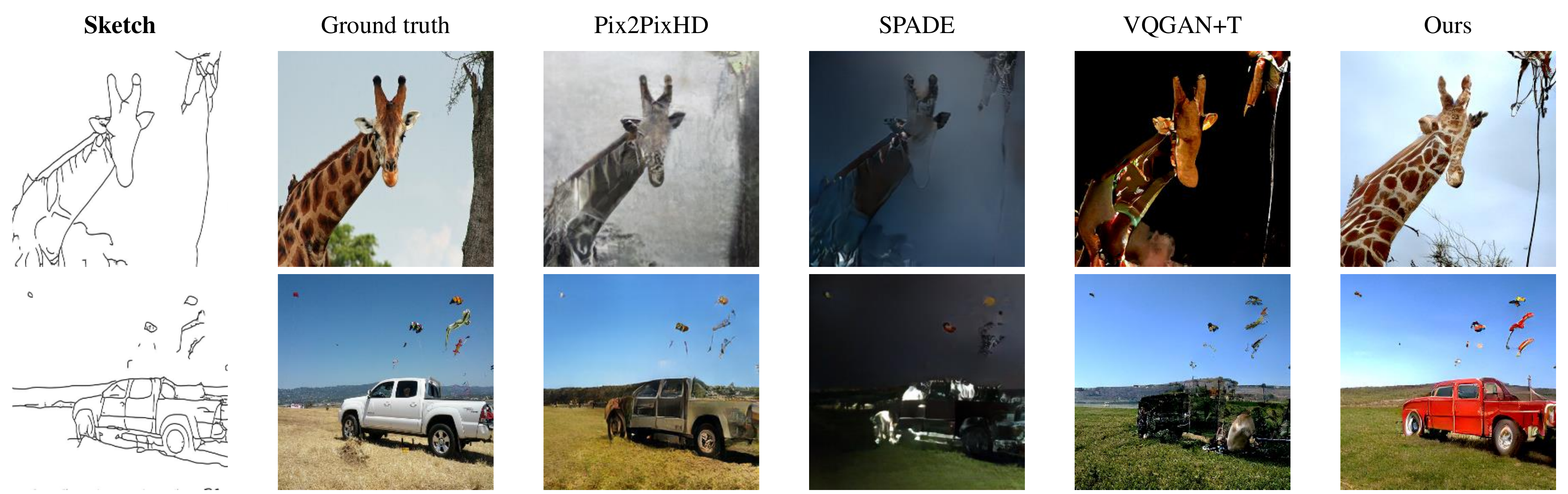}
    \caption{Qualitative comparison of sketch-to-image synthesis results on COCO-Stuff. More results are demonstrated in~\ref{sec:qualitative-ucis}}
    \label{fig:sketch-to-image}
\end{figure*}

We compare MMoT with one of the state-of-the-art MCIS methods PoE-GAN~\citep{huang2022multimodal} and also with a wide range of UCIS approaches in the unimodal setting. Since M6-UFC and Composer \citep{zhang2021ufc, huang2023composer} are performed on different MCIS datasets and their codes are unavailable till our submission, we did not make direct comparisons with them.

\subsubsection{Text to image synthesis}\label{sec:t2i}
Text-to-Image is designed to render a realistic image from a text description, which is a rather challenging task that dominates image generation.
It is also a cross-modal generation task, which requires the model to be able to generate images that meet people's expectations based on understanding the objects and their relationships described in the text.
For text-to-image synthesis, we compare with DF-GAN~\citep{tao2020df}, DM-GAN+CL~\citep{ye2021improving}, VQGAN+T~\citep{esser2021taming} and PoE-GAN~\citep{huang2022multimodal} on COCO-Stuff. Since text annotation is not available in LHQ, we compare with VQGAN+T~\citep{esser2021taming}, MaskGIT~\citep{chang2022maskgit}, and NUWA-Infinity~\citep{wu2022nuwaif} in the language-free setting, \textit{i.e.}, language annotations are unavailable and we take the pseudo-labelling texts as condition.
The quantitative results of text-to-image synthesis on COCO-Stuff and LHQ are reported in Table~\ref{tab:main_coco} (a) and Table~\ref{tab:main_lhq} (a), respectively. In Figure~\ref{fig:text-to-image}, we also provide several qualitative comparisons. MMoT has comparable performance, both quantitatively and qualitatively.

\subsubsection{Bounding boxes to image synthesis}\label{sec:b2i}
Bounding boxes to image generation aims to generate photo-realistic conditioned on specified layouts which consists of a set of object bounding boxes and corresponding categories. Compared with text-to-image synthesis, such a layout provides a simple sketch of the image, which makes the generation more user-friendly and controllable, but this also reduces the diversity of generated images to some extent.
For bounding-boxes-to-image generation, we compare with LostGAN-V2~\citep{sun2021learning}, OC-GAN~\citep{sylvain2021object}, VQGAN+T~\citep{esser2021taming}, LAMA~\citep{li2021image}, Context-L2I~\citep{he2021context} and TwFA~\citep{yang2022modeling}. 
The performance of several methods for bounding boxes to image synthesis on COCO-Stuff is evaluated quantitatively and qualitatively. The quantitative results are reported in Table~\ref{tab:main_coco} (b), while several qualitative comparisons are shown in Figure~\ref{fig:layout-to-image}. The evaluations demonstrate that MMoT achieves better performance compared to the other methods.

\subsubsection{Segmentation to image synthesis}\label{sec:s2i}
The goal of segmentation to image synthesis is to generate a full-color image from a grayscale segmentation mask, where each pixel in the mask corresponds to a specific object or region in the image.
For segmentation masks-to-image synthesis, we compare with pix2pixHD~\citep{wang2018high}, SPADE~\citep{park2019semantic}, VQGAN+T~\citep{esser2021taming}, OASIS~\citep{sushko2022oasis}, PITI~\citep{wang2022pretraining} and PoE-GAN~\citep{huang2022multimodal}.
Table~\ref{tab:main_coco} (c) and Table~\ref{tab:main_lhq} (b) present the quantitative results of segmentation-to-image synthesis on COCO-Stuff and LHQ, respectively. Figure~\ref{fig:seg-to-image} provides several qualitative comparisons. Both the quantitative and qualitative evaluations show that MMoT outperforms the other methods.

\subsubsection{Sketch to image synthesis}\label{sec:k2i}
The same as segmentation to image synthesis, converting a sketch to an image is also an image-to-image translation task that involves creating an image from a rough, hand-drawn sketch or line drawing. This task can be challenging because sketches often lack detail, texture, and color information, and require a model to infer these missing details to generate a realistic image.
For sketch-to-image generation, we compare with pix2pixHD~\citep{wang2018high}, SPADE~\citep{park2019semantic}, VQGAN+T~\citep{esser2021taming}, PITI~\citep{wang2022pretraining} and PoE-GAN~\citep{huang2022multimodal}.
Table~\ref{tab:main_coco} (d) and Table~\ref{tab:main_lhq} (c) show the quantitative results of sketch-to-image synthesis on COCO-Stuff and LHQ, respectively. Additionally, Figure~\ref{fig:sketch-to-image} presents several qualitative comparisons. The evaluations indicate that MMoT performs better than the other methods, both in terms of quantitative metrics and qualitative comparisons.

Sect.~\ref{sec:t2i} to Sect.\ref{sec:b2i} demonstrates MMoT's unimodal conditional image generation capabilities. Surprisingly, MMoT achieves comparable performance with unimodal methods specifically designed for that modality on both datasets. 
In Figure~\ref{fig:text-to-image} to Figure~\ref{fig:sketch-to-image}, we found that MMoT is robust to different input modalities and can produce photo-realistic images of refiner textures (\textit{e.g.}, pasta, bus, and building), clearer structures (\textit{e.g.}, giraffe, bear, and car), and more reasonable interactions (\textit{e.g.}, the reflection on the bus windows, the car’s shadow on the ground). It is worth noting that, unlike unimodal conditional synthesis models, MMoT supports the combination of many different types of inputs.

\subsubsection{Multimodal conditional image synthesis}
As mentioned earlier, unimodal conditional image generation supports only one type of conditioning information. To make the generation more flexible and controllable, multimodal conditional image generation can synthesize images conditioned on multiple types of modality inputs.
As illustrated in Table~\ref{tab:main_coco} (e), we also obtain better results than PoE-GAN when conditioned on All$^\dagger$ (\textit{i.e.}, text + segmentation masks + sketch). It is worth noting that, since PoEGAN is trained with only three modalities, we also test our MMoT on these modalities. It could be a little bit unfair but has some reference significance.

Overall, when conditioned on a single modality, the superiority of the proposed MMoT is validated on both quantitative metrics and qualitative visual comparison. The improved performance of unimodal conditional image synthesis indicates the robustness of the multistage token-mixer and the balanced optimization of each modality. In addition, MMoT also outperforms previous state-of-art MCIS method PoE-GAN when conditioned on multimodal conditional inputs.

\subsection{Image Synthesis Conditioned on the Compositions of Inputs}\label{sec:multimodal samples}
\begin{figure*}
    \centering
    \includegraphics[width=\textwidth]{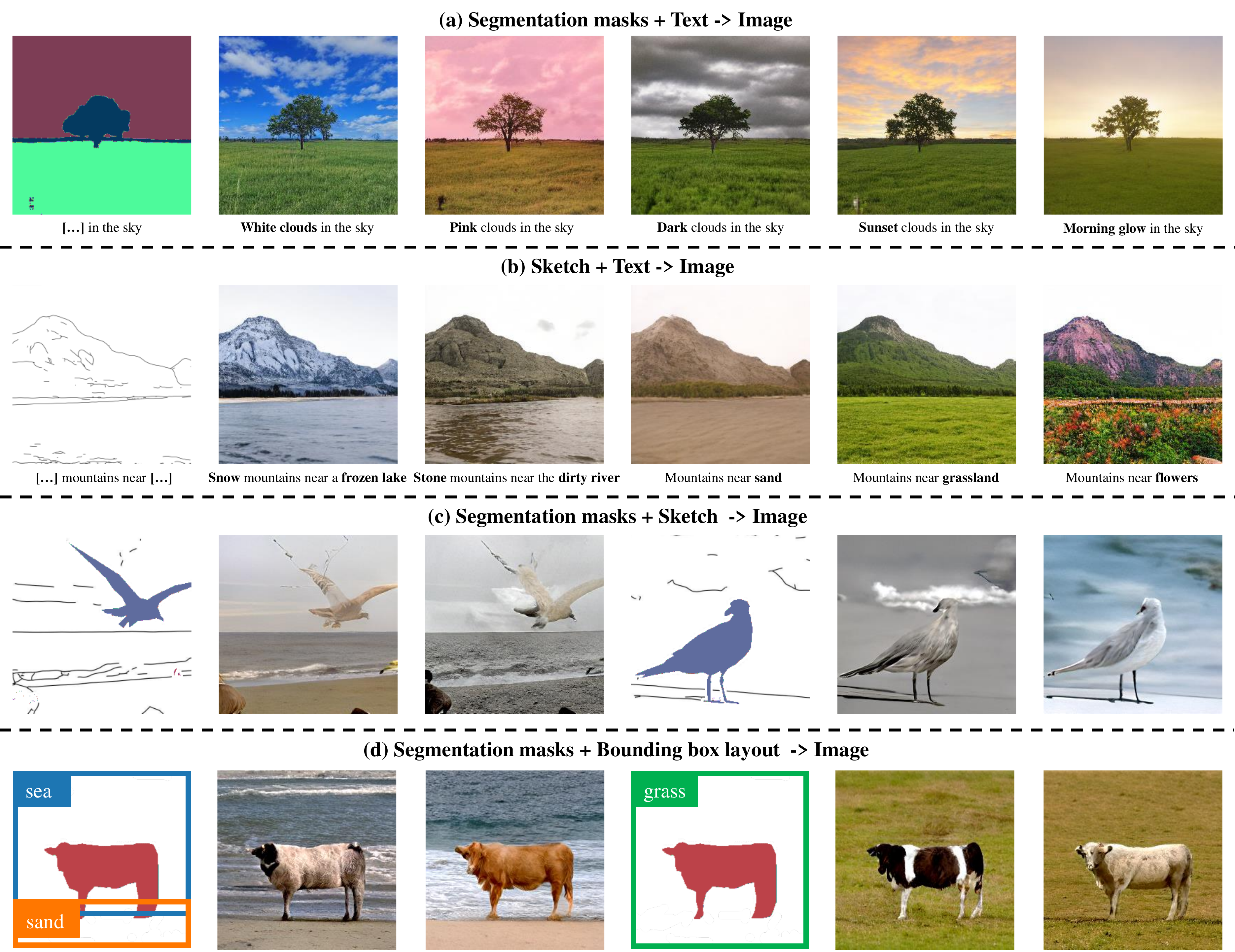}
    \caption{Samples of composed multimodal conditional image synthesis generated by MMoT. We show compositions of different modalities: (a) segmentation masks and text; (b) sketch and text; (c) segmentation masks and sketch; (d) segmentation masks and bounding box layout. MMoT can synthesize reasonable images leveraging conditional information conveyed by different modalities. More results are demonstrated in~\ref{sec:qualitative-mcis}}
    \label{fig:multimodal_samples}
\end{figure*}
The most exciting ability of MMoT is that it can synthesize imagery images according to the compositions of a set of input modalities.

In Figure~\ref{fig:multimodal_samples}, we show some multimodal conditional samples generated by our MMoT. As illustrated in Figure~\ref{fig:multimodal_samples} (a), using segmentation masks to give the coarse layout of semantic classes (\textit{e.g.}, sky, trees, and grass) is easy but it is impossible to specify the color of an object, however, with the help of text, the color of clouds can be defined. Similarly, sketch allow us to describe the shape (\textit{e.g.}, ridges of mountains) and texture (\textit{e.g.}, ripples of water) of an object with simple strokes, but only under the control of the text can the state (\textit{e.g.}, snow or stone mountain, frozen lake or dirty river) and category (\textit{e.g.}, river, sand or grassland) of the object be given. Moreover, as shown in Figure~\ref{fig:multimodal_samples} (c) and (d), MMoT can synthesize realistic and diverse images when conditioned on the compositions of segmentation masks and sketch/bounding box layout.

\subsection{Ablation Study}\label{sec: ablations}
\begin{table*}[t]
\begin{center}
\begin{minipage}{\linewidth}
\caption{Ablation on COCO-Stuff (256$\times$256)}\label{tab:ablation_coco}
\resizebox{\textwidth}{9.5mm}{
    \begin{tabular}{lcccccccccc}
        \toprule
                        & \multicolumn{2}{c}{Text}                  & \multicolumn{2}{c}{Bounding boxes}            & \multicolumn{2}{c}{Segmentation masks}          & \multicolumn{2}{c}{Sketch}                & \multicolumn{2}{c}{All} \\ 
        \cmidrule(lr){2-3} \cmidrule(lr){4-5} \cmidrule(lr){6-7} \cmidrule(lr){8-9} \cmidrule(lr){10-11}
        \textbf{Methods} & FID $\downarrow$ & Clean-FID $\downarrow$ & FID $\downarrow$ & Inception Score $\uparrow$ & FID $\downarrow$ & Clean-FID $\downarrow$ & FID $\downarrow$ & Clean-FID $\downarrow$ & FID $\downarrow$ & Clean-FID $\downarrow$ \\
        Base            & 55.93            & 56.27                  & 26.27            & 22.50±0.67                 & 20.07            & 20.43                  & 32.91            & 33.73                  & 13.43            & 13.68 \\
        + Multistage Token-Mixer & 31.19            & 31.40                  & 21.36            & 24.18±0.59                 & 14.52            & 14.77                  & 23.75            & 24.37                  & 12.79            & 13.04 \\
        + Multimodal Balanced Loss & \underline{23.04}            & \underline{23.32}      & \underline{20.65} & \underline{24.90±0.57}    & \underline{13.24} & \underline{13.43}     & \textbf{23.08}   & \textbf{23.70}         & \underline{12.23} & \underline{12.44} \\
        + Multimodal Guidance & \textbf{17.91} & \textbf{17.83}                 & \textbf{19.24}   & \textbf{26.67±0.50}        & \textbf{12.73}   & \textbf{12.91}         & \underline{23.42} & \underline{23.93}     & \textbf{11.75}   & \textbf{11.73} \\
        \bottomrule
    \end{tabular}}
{The best scores are highlighted in bold and the second best ones are underlined.}
\end{minipage}
\end{center}
\end{table*}

In Table~\ref{tab:ablation_coco}, we analyze the importance of different components of MMoT. The settings in both studies are similar to the comparison with existing methods in Table~\ref{tab:main_coco}. The baseline model is the vanilla encoder-decoder transformer, which simply concatenates the features outputted by respective encoders while performing modality dropout. Each row corresponds to a model trained with the additional element.

Compared with the baseline, when adding the multistage token-mixer, the performances of unimodal conditions and multiple conditions have gained significant improvement, which indicates the robustness on missing-modality and the effectiveness of the fusion of the multistage token-mixer. And since multimodal balanced loss can facilitate the optimization of each modality, unimodal conditional image synthesis achieved further improvement, especially for text-to-image generation. We note that multimodal guidance is useful for text-to-image synthesis but not essential for each input condition.

\subsection{Qualitative Analysis}\label{sec: analysis}
\begin{figure}
    \centering
    \includegraphics[width=0.45\textwidth]{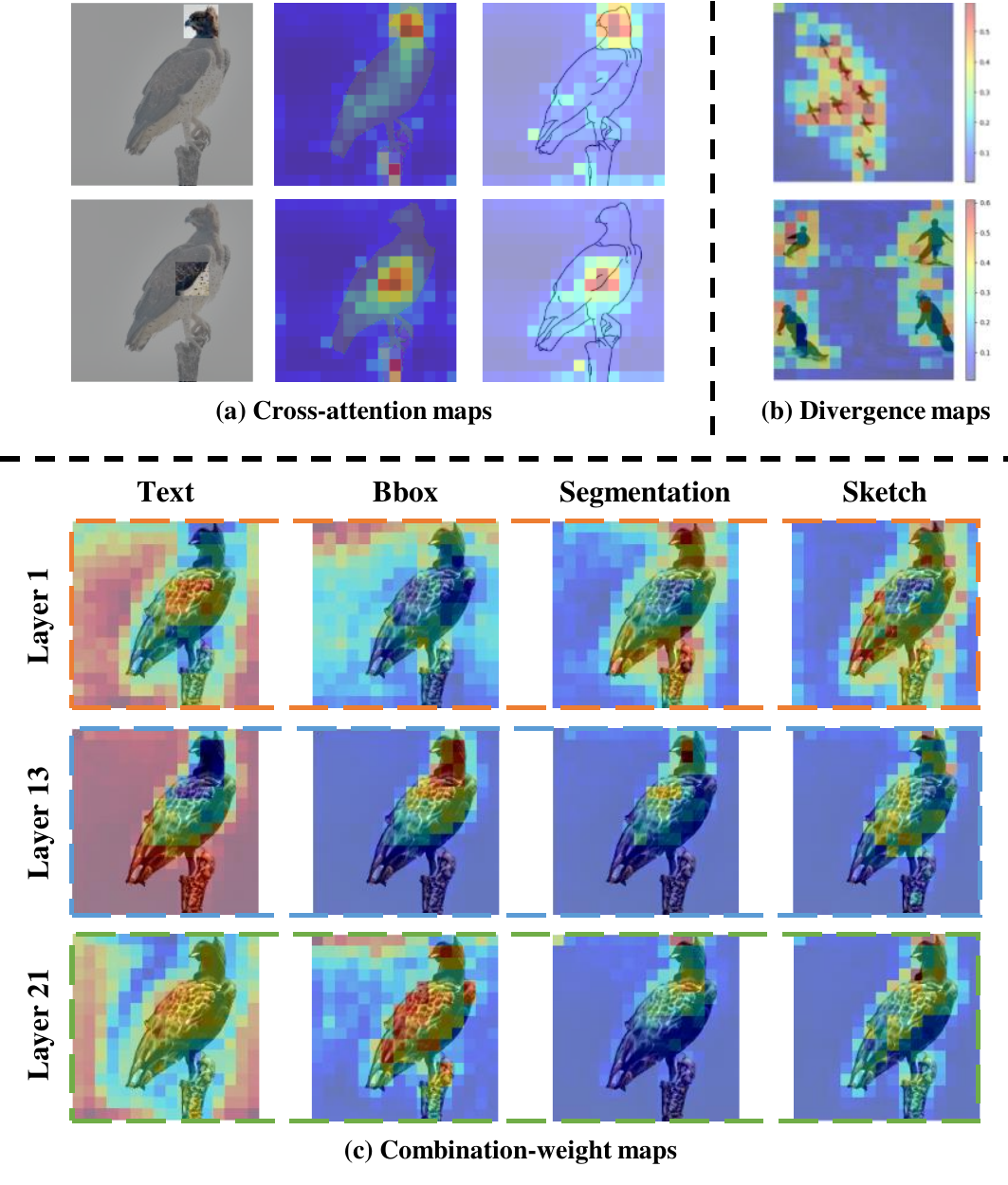}
    \caption{Visualizations of cross attention maps, divergence maps, and combination weight maps.}
    \label{fig:analysis}
\end{figure}
In Figure~\ref{fig:analysis}, we show some visualizations of the underlying process, including cross-attention maps, divergence maps, and combination-weight maps.
\subsubsection{Cross-attention maps}
Part (a) in Figure~\ref{fig:analysis} is the average attention score maps across all transformer decoder layers. We show the cross-attention map conditioned on segmentation masks and sketches. It is noted that when generating the specific areas of an image (\textit{e.g.}, the head or thorax of a bird), high responsiveness is observed in the corresponding areas of the segmentation map or sketch. 
\subsubsection{Divergence map}
Part (b) in Figure~\ref{fig:analysis} is the distribution of the JS divergence calculated between unimodal conditional logits and unconditional logits over different generated images, which contain rich semantic relations reflecting the influence of different conditions. So injecting the divergence map as composition prior to controlling the value of the guidance scale during the sampling process can lead to more spatially coordinated images.
\subsubsection{Combination-weight maps}
 The key to effective fusion is to calculate accurate combination weights, which represent the influences of each input modality. The combination weights in MMoT are related to attention scores calculated in Eq.~(\ref{eq:mixer_function}).
Part (c) in Figure~\ref{fig:analysis} shows the combination-weight maps of individual modalities in layers 1, 13, and 21, which show that the modality at different semantic levels contributes to different layers, \textit{i.e.}, text and bbox have more contributions in the higher layer, while segmentation and sketch play a role in all layers, especially in the earlier layer.
It also illustrates that the special token \texttt{[PULSE]} within the multistage token-mixer can adaptively detect the influences of modality tokens across the generated image.

\section{Conclusions}
In this paper, we focus on the challenging Composed Multimodal Conditional Image Synthesis (CMCIS) task and propose a novel Mixture-of-Modality-Tokens Transformer (MMoT).
Towards two severe issues of CMCIS, \textit{i.e.}, modality coordination and imbalance problems, we introduce a multistage token-mixer, multimodal balanced loss, and divergence-driven sampling guidance to fully exploit the cooperativity across different modalities.
Extensive experiments on the COCO-Stuff and LHQ datasets demonstrate that the proposed MMoT successfully generates high-quality and faithful images conditioned on composed multimodal signals, and achieves superior performance over most existing UCIS and MCIS models.

\noindent\textbf{Limitations.}
Since our framework is based on the autoregressive Transformer, it suffers from limited inference speed. In the future, we will attempt to adapt the proposed module and training/testing schemes to other deep neural frameworks and hope to produce more realistic, high-quality, and diverse results. 

\noindent\textbf{Broader Impacts.}
The proposed composed multimodal image synthesis offers unprecedented fine-grained generation capabilities, which lead to both positive and negative societal impacts. Multimodal control signals as input for synthesis greatly improve the flexibility of user interaction and ease the use of deep generative models. However, the increasing generation capabilities also make it easier to synthesize desired images for malicious purposes, \textit{i.e.}, the misuse of fake or nefarious information.
In the future, sufficient guardrails, access control, and detection techniques are encouraged to minimize the risk of misuse.

\backmatter

\begin{appendices}
\section{More Qualitative Results}\label{sec:qualitative}
\subsection{Qualitative Comparisons with UCIS Models}\label{sec:qualitative-ucis}
\begin{figure*}
	\centering
	\includegraphics[width=0.95\textwidth]{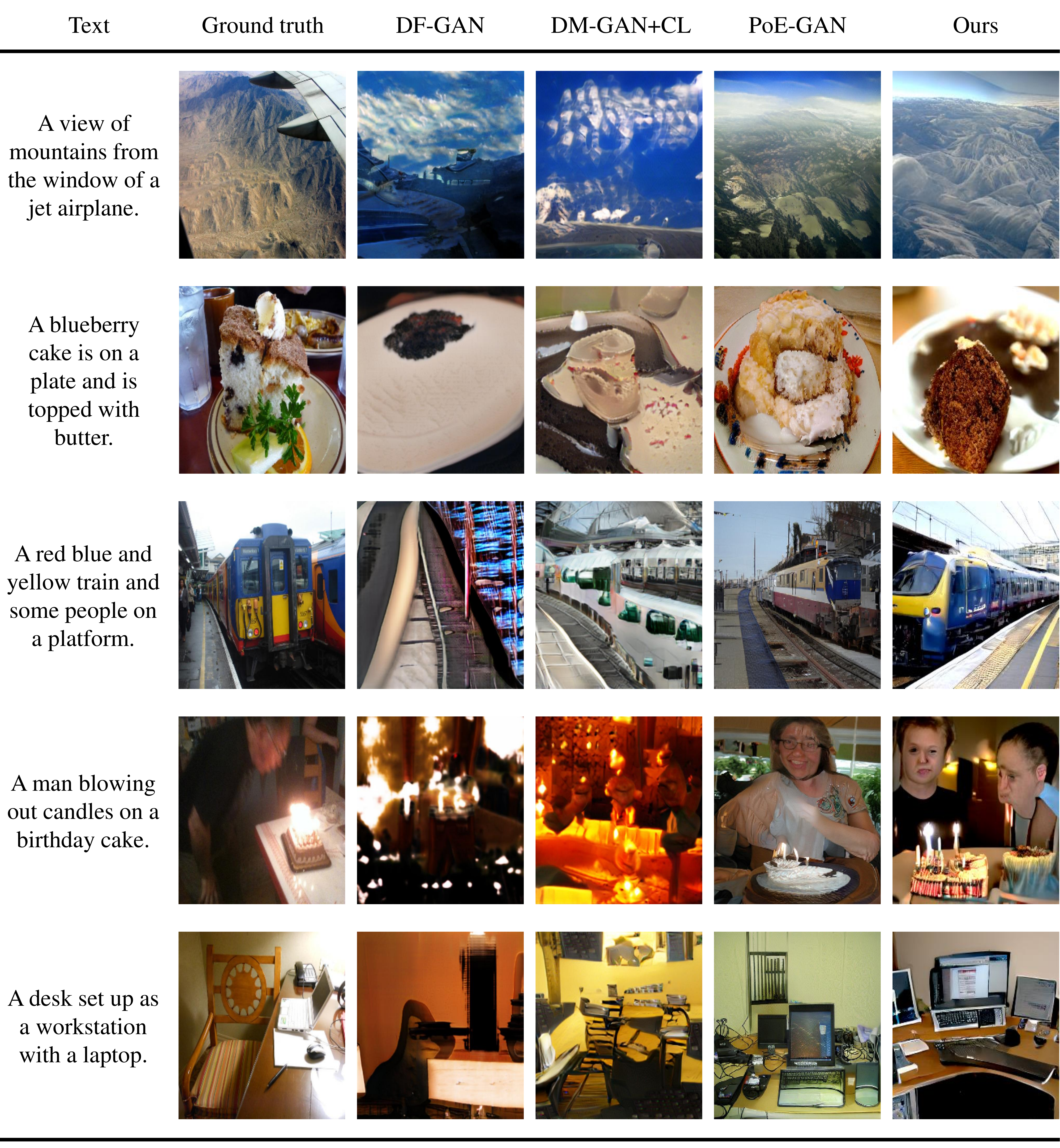}
	\caption{Additional qualitative comparison of text-to-image synthesis on COCO-Stuff.}
	\label{fig:t2i}
\end{figure*}

\begin{figure*}
	\centering
	\includegraphics[width=0.95\textwidth]{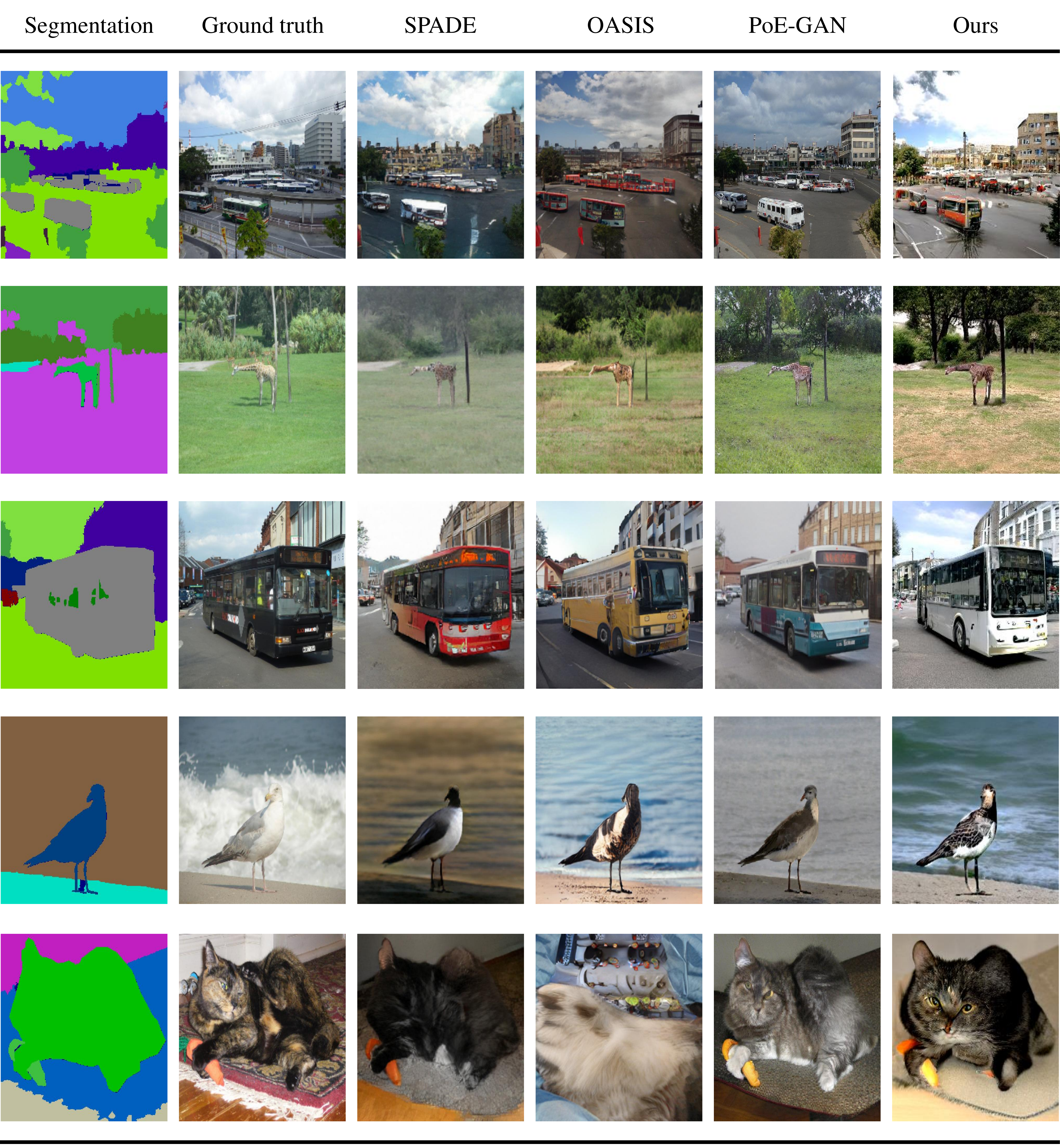}
	\caption{Additional qualitative comparison of segmentation mask-to-image synthesis on COCO-Stuff.}
	\label{fig:s2i}
\end{figure*}

\begin{figure*}
	\centering
	\includegraphics[width=0.95\textwidth]{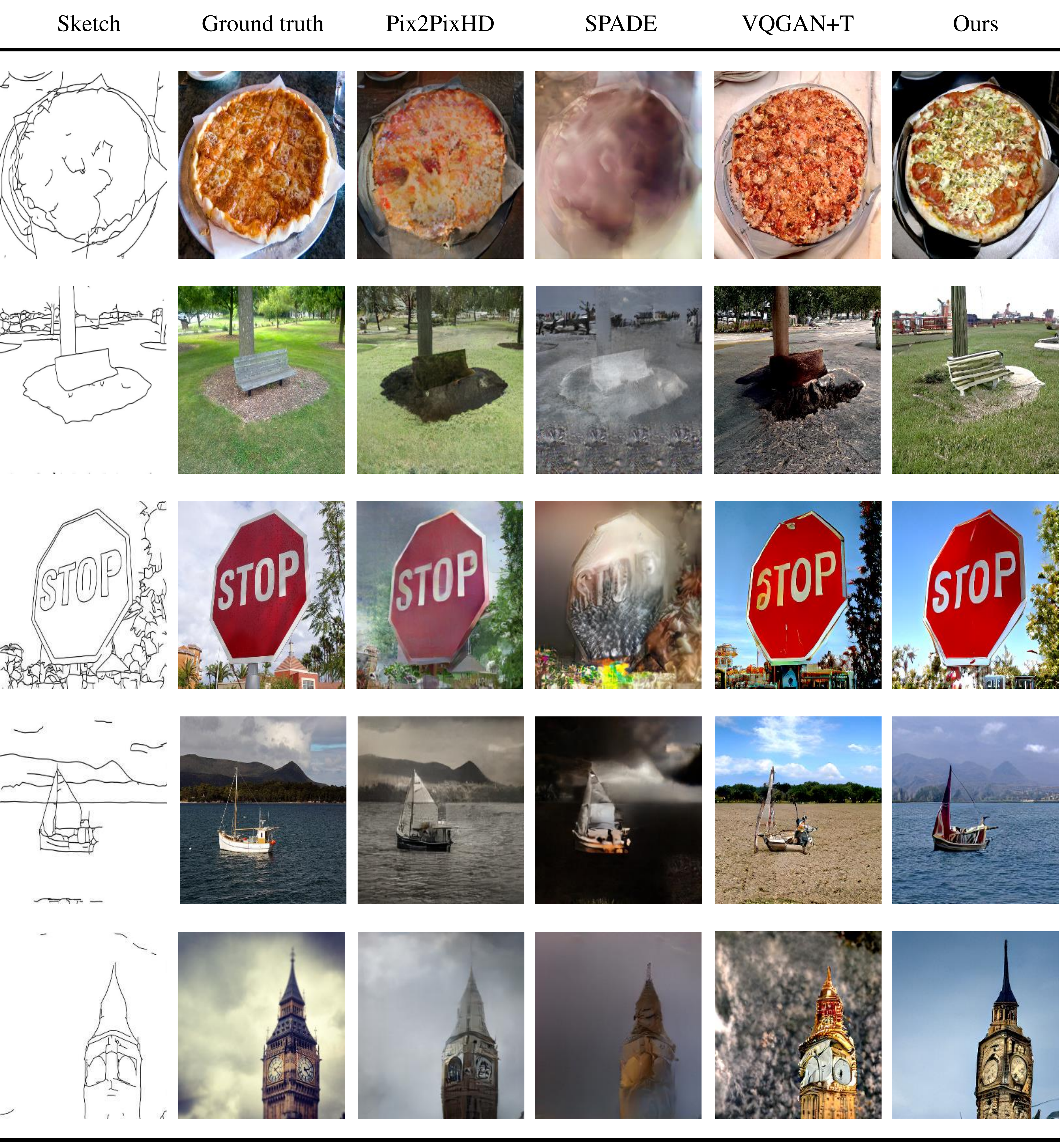}
	\caption{Additional qualitative comparison of sketch-to-image synthesis on COCO-Stuff.}
	\label{fig:k2i}
\end{figure*}

\begin{figure*}
	\centering
	\includegraphics[width=0.95\textwidth]{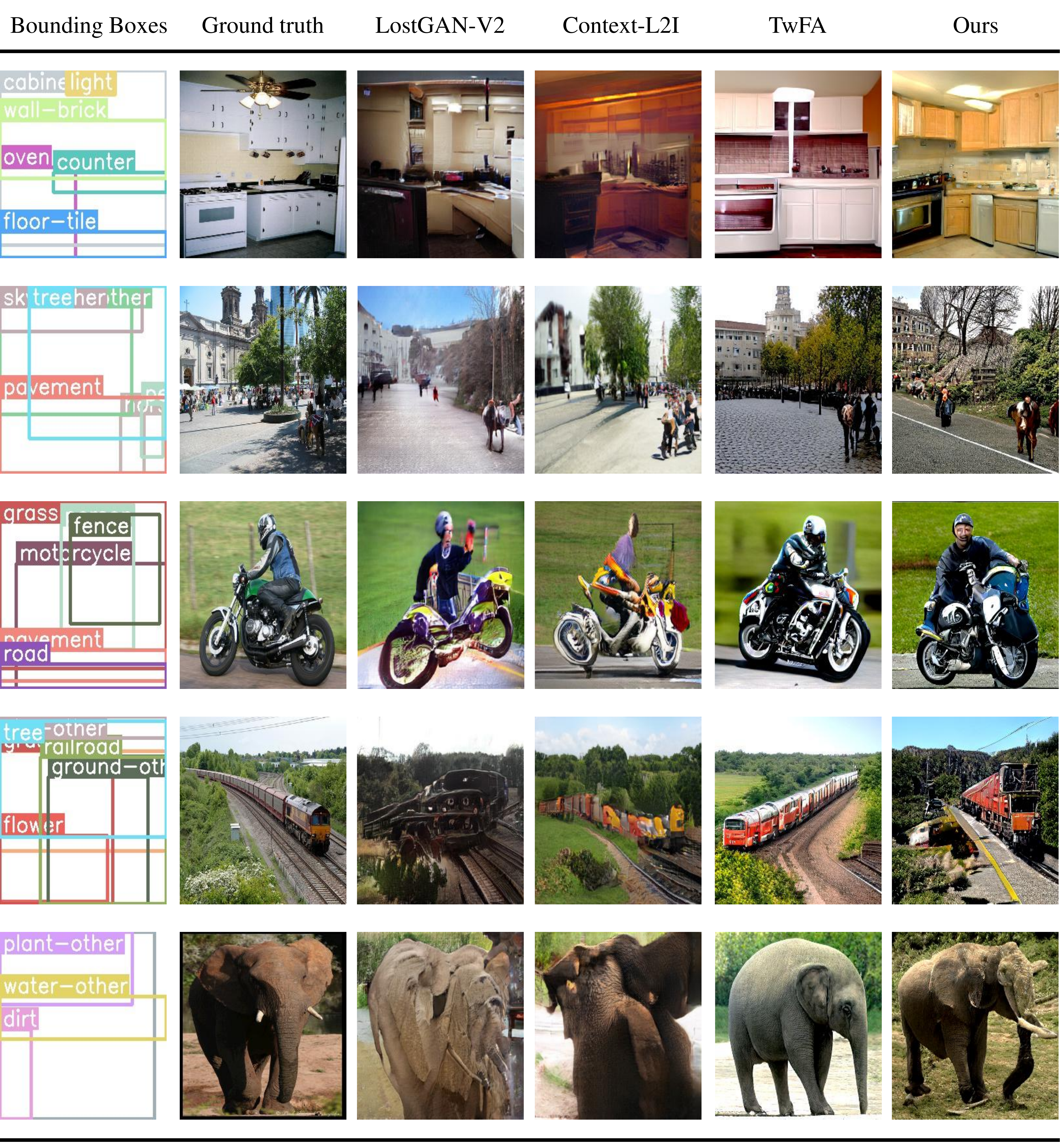}
	\caption{Additional qualitative comparison of bounding boxes-to-image synthesis on COCO-Stuff.}
	\label{fig:b2i}
\end{figure*}

In Figure~\ref{fig:t2i} to \ref{fig:b2i}, we show additional qualitative comparisons with a wide range of UCIS models when conditioned on text, segmentation mask, sketch, and bounding boxes, respectively. Competitive visual results compared with UCIS models specially designed for a single modality indicate MMoT is robust to different modalities.

\subsection{Qualitative CMCIS Examples}\label{sec:qualitative-mcis}
\begin{figure*}
	\centering
	\includegraphics[width=0.95\textwidth]{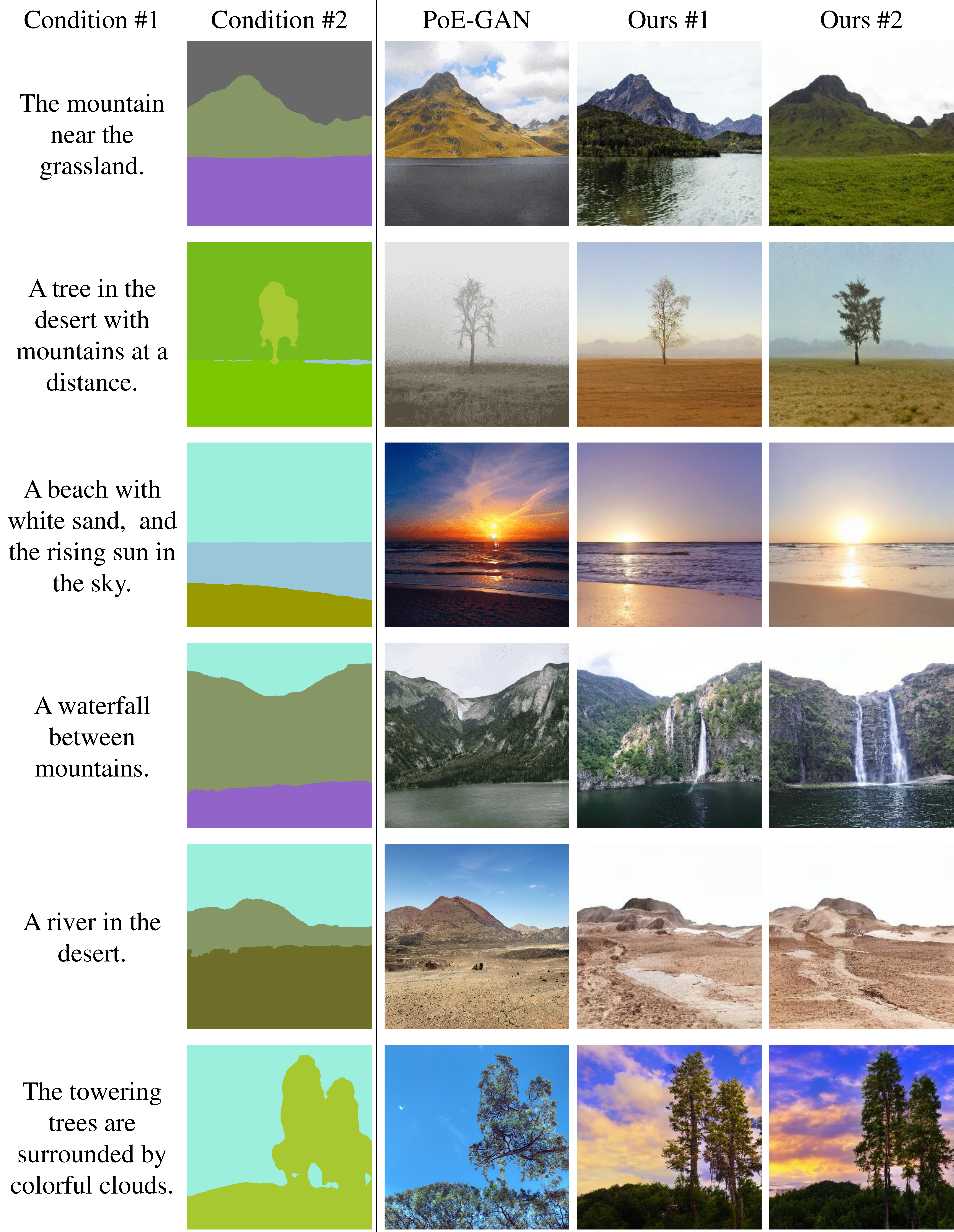}
	\caption{Examples of composed multimodal conditional image synthesis when conditioned on text and segmentation mask. From left to right: text, segmentation, a random sample from PoE-GAN, and two random samples from our MMoT. PoE-GAN always struggles with the modality imbalance problem. In contrast, MMoT can balance the information of the two modalities to synthesize images.}
	\label{fig:ts2i}
\end{figure*}

\begin{figure*}
	\centering
	\includegraphics[width=0.95\textwidth]{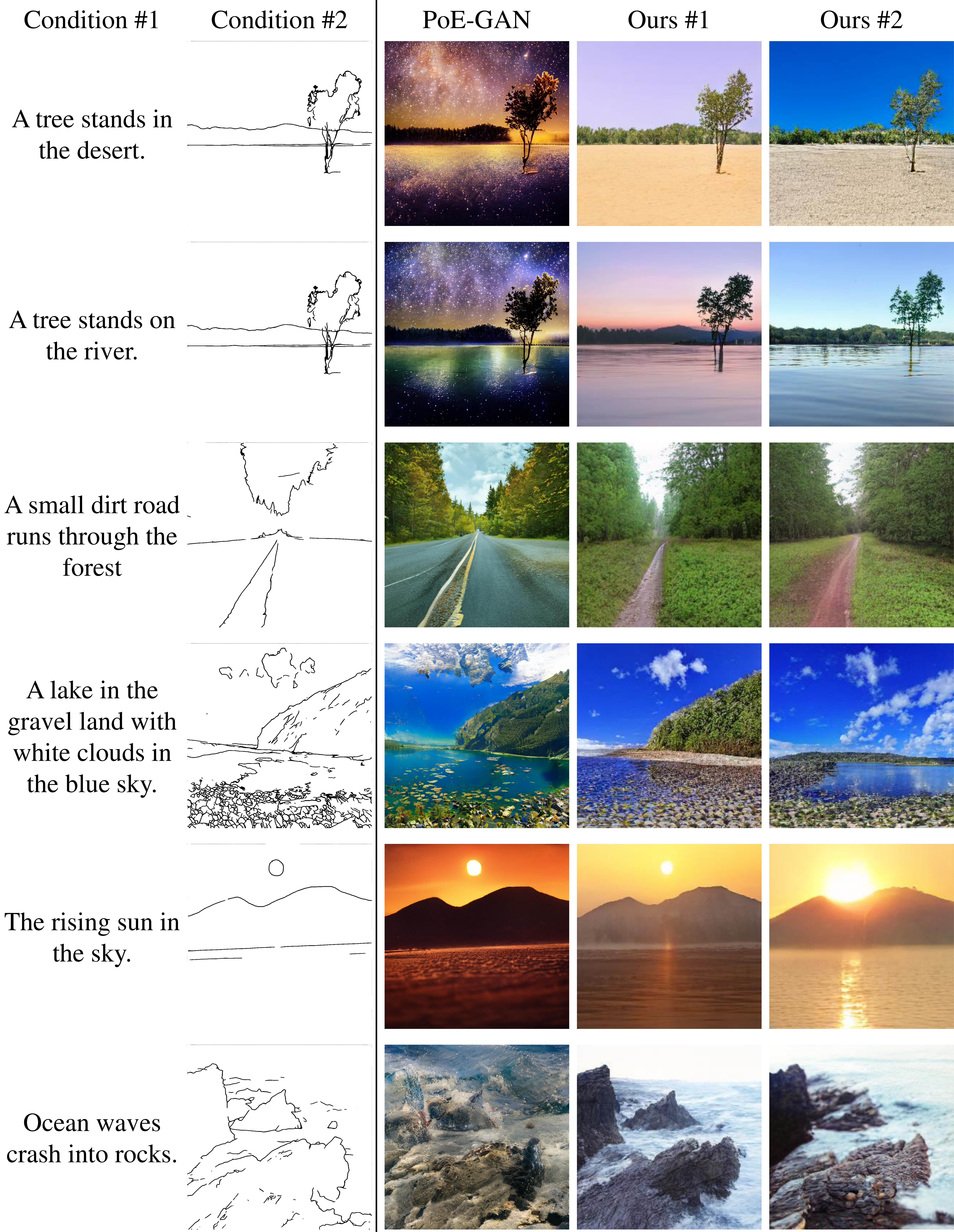}
	\caption{Examples of composed multimodal conditional image synthesis when conditioned on text and sketch. From left to right: text, sketch, a random sample from PoE-GAN, and two random samples from our MMoT. PoE-GAN always struggles with the modality imbalance problem. In contrast, MMoT can balance the information of the two modalities to synthesize images.}
	\label{fig:tk2i}
\end{figure*}

\begin{figure*}
	\centering
	\includegraphics[width=0.95\textwidth]{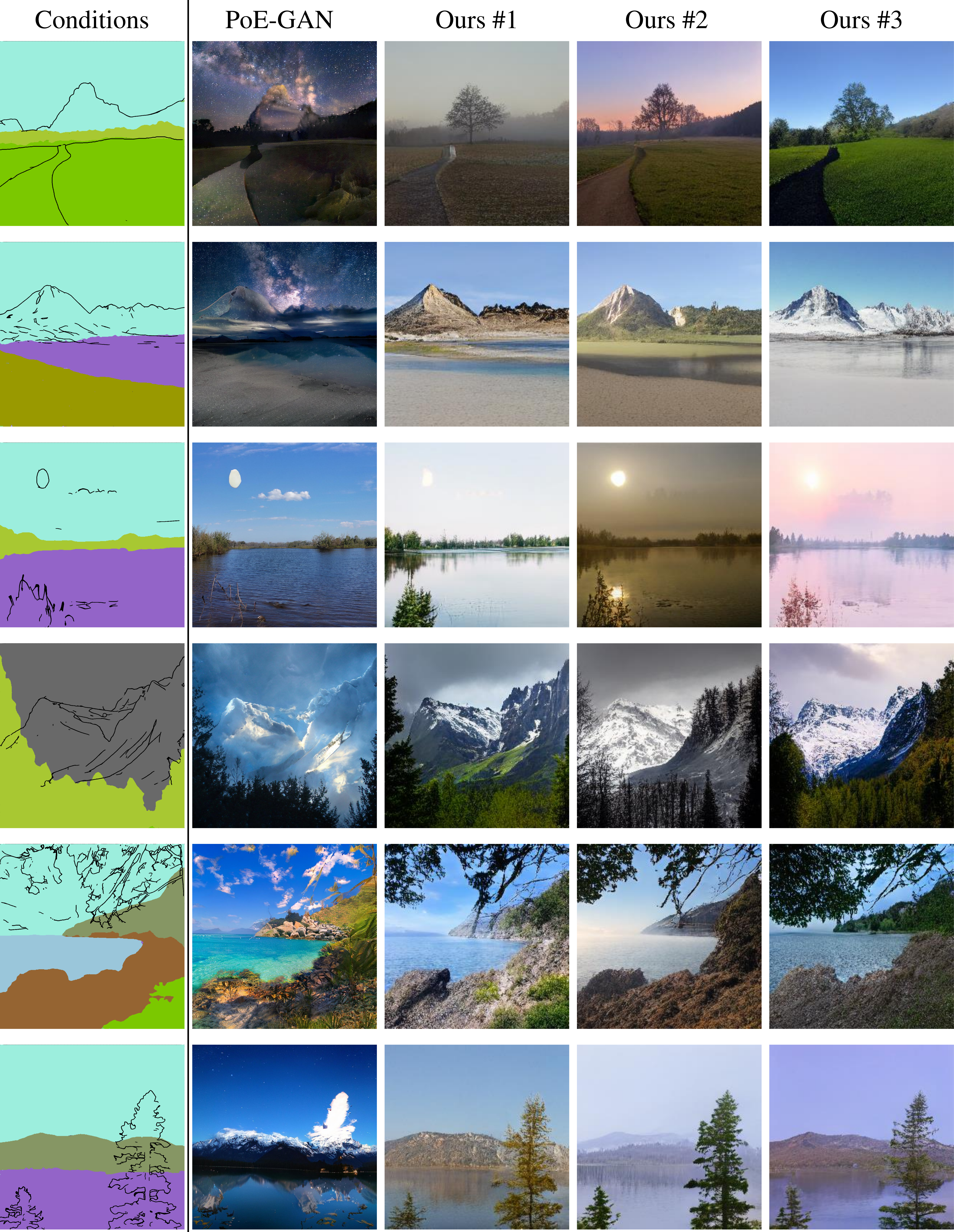}
	\caption{Examples of composed multimodal conditional image synthesis when conditioned on segmentation and sketch. From left to right: segmentation mask, sketch, a random sample from PoE-GAN, and three random samples from our MMoT. PoE-GAN always struggles with the modality coordination problem. In contrast, MMoT can generate more spatially coordinated images.}
	\label{fig:sk2i}
\end{figure*}

\begin{figure*}
	\centering
	\includegraphics[width=0.95\textwidth]{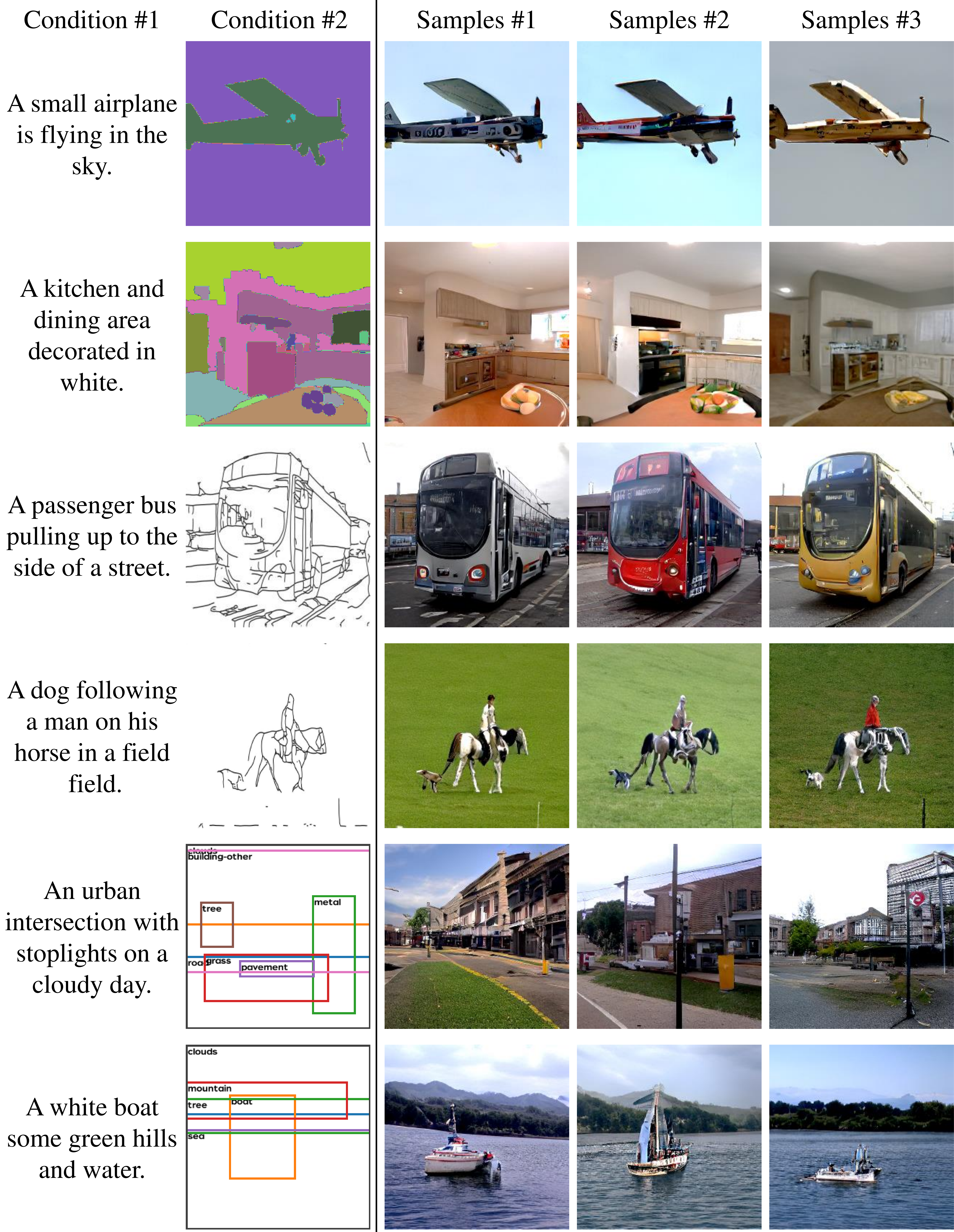}
	\caption{Examples of composed multimodal conditional image synthesis. We show three random samples from MMoT conditioned on compositions of different modalities (from top to bottom: text+segmentation mask, text+sketch, and text+bounding boxes).}
	\label{fig:coco_1}
\end{figure*}

\begin{figure*}
	\centering
	\includegraphics[width=0.95\textwidth]{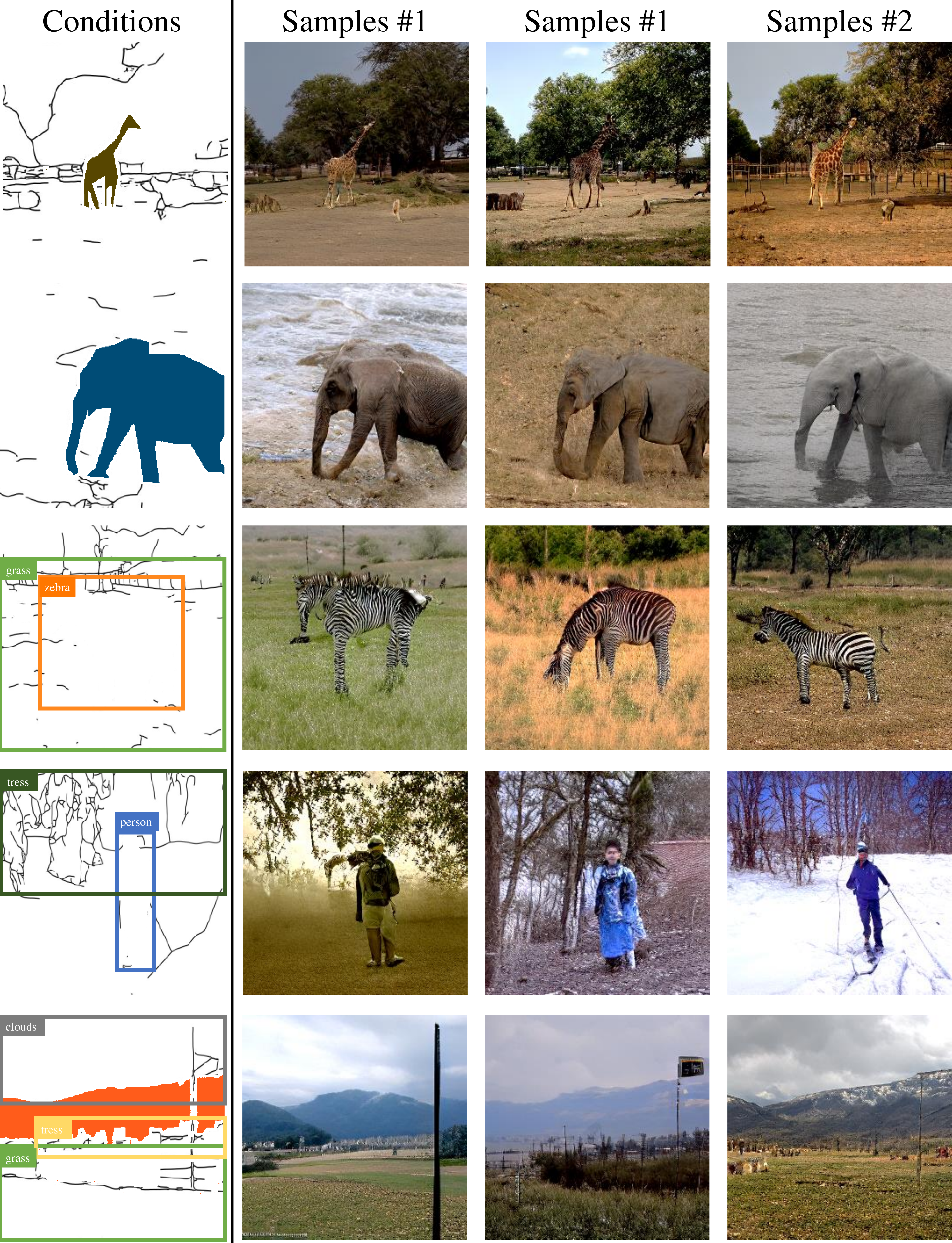}
	\caption{Examples of composed multimodal conditional image synthesis. We show three random samples from MMoT conditioned on compositions of different modalities (from top to bottom: segmentation mask+sketch, bounding boxes+sketch, and segmentation mask+sketch+bounding boxes).}
	\label{fig:coco_2}
\end{figure*}

Figure~\ref{fig:ts2i} to \ref{fig:coco_2} show that MMoT can generate high-quality, faithful, and diverse images when conditioned on complex compositions of two or three different modalities.

In Figure~\ref{fig:ts2i} to \ref{fig:sk2i}, we also show more visual comparisons with PoE-GAN when conditioned on compositions of text+segmentation mask, text+sketch, and segmentation mask+sketch, respectively. The modality coordination problem and the modality imbalance problem are common in MCIS models when conditioned on complex multimodal conditions. In contrast, MMoT addresses both issues and can synthesize high-quality and faithful images.

\noindent\textbf{Modality coordination problem.} 
The modality coordination problem is caused by the nonadaptive fusion of fine-grained information across multiple modalities.
As illustrated in Figure~\ref{fig:sk2i}, when PoE-GAN synthesizes an image, the generated contents from the sketch condition are incorrectly composed with the generated contents from the segmentation mask condition.

\noindent\textbf{Modality imbalance problem.} 
The modality imbalance problem is caused by the imbalanced distribution of each modality in datasets.
As illustrated in Figure~\ref{fig:ts2i} and \ref{fig:tk2i}, PoE-GAN tends to ignore text inputs when generating images.




\end{appendices}

\clearpage
\bibliography{references}


\end{document}